\pdfoutput=1

\documentclass[11pt]{article}

\usepackage{emnlp2021}
\usepackage{emnlp2021}

\usepackage{times}
\usepackage{latexsym}
\usepackage{graphicx}
\usepackage{subfigure}

\usepackage[T1]{fontenc}

\usepackage[utf8]{inputenc}

\usepackage{microtype}

\usepackage{array}
\usepackage{pifont}
\usepackage{tabularx}
\usepackage{adjustbox}
\usepackage{multirow}
\usepackage{enumitem}
\usepackage{xspace}
\usepackage{tcolorbox}
\usepackage{booktabs,amsfonts,dcolumn}
\usepackage{hyperref}
\usepackage{url}
\usepackage{amsmath,amsthm,amsfonts,amssymb,bm,stmaryrd,bbm}
\usepackage[noorphans,vskip=0.75ex,leftmargin=2ex]{quoting}

\providecommand{\tianyu}[1]{
    {\protect\color{blue}{[Tianyu: #1]}}
}

\providecommand{\xingcheng}[1]{
    {\protect\color{teal}{[Xingcheng: #1]}}
}

\newcommand\ti[1]{\textit{#1}}

\newcommand\tf[1]{\textbf{#1}}
\newcommand\ttt[1]{\texttt{#1}}
\newcommand\mf[1]{\mathbf{#1}}

\newcommand\mr[1]{\mathrm{#1}}

\newcommand{\cls}{\ttt{[CLS]}}

\newcommand{\ours}{SimCSE\xspace}

\newcommand{\la}{$_\texttt{large}$}
\newcommand{\ba}{$_\texttt{base}$}

\renewcommand{\paragraph}[1]{\vspace{0.2cm}\noindent\textbf{#1}}

\newcommand{\tableindent}{~~}

\title{\ours: Simple Contrastive Learning of Sentence Embeddings}

\author{Tianyu Gao$^{\dagger*}$ \quad Xingcheng Yao$^{\ddagger*}$ \quad Danqi Chen$^{\dagger}$ \\
$^{\dagger}$Department of Computer Science, Princeton University\\$^{\ddagger}$Institute for Interdisciplinary Information Sciences, Tsinghua University\\
\ttt{\{tianyug,danqic\}@cs.princeton.edu}\\
\ttt{yxc18@mails.tsinghua.edu.cn}
}

\begin{document}
\maketitle
\renewcommand{\thefootnote}{\fnsymbol{footnote}}
\footnotetext[1]{The first two authors contributed equally (listed in alphabetical order). This work was done when Xingcheng visited the Princeton NLP group remotely.}
\renewcommand{\thefootnote}{\arabic{footnote}}

\begin{abstract}



This paper presents {\ours}, a simple contrastive learning framework that greatly advances state-of-the-art sentence embeddings.
We first describe an unsupervised approach, which takes an input sentence and predicts \ti{itself} in a contrastive objective, with only standard dropout used as noise. This simple method works surprisingly well, performing on par with previous supervised counterparts. We find that dropout acts as minimal data augmentation, and removing it leads to a representation collapse. Then, we propose a supervised approach, which incorporates annotated pairs from natural language inference datasets into our contrastive learning framework by using ``entailment'' pairs as positives and ``contradiction'' pairs as hard negatives.
We evaluate {\ours} on standard semantic textual similarity (STS) tasks,
and our unsupervised and supervised models using BERT\ba~achieve an average of
76.3\% and 81.6\% Spearman’s correlation respectively,
a 4.2\% and 2.2\% improvement compared to the previous best results.
We also show---both theoretically and empirically---that the contrastive learning objective regularizes pre-trained embeddings' anisotropic space to be more uniform, and it better aligns positive pairs when supervised signals are available.\footnote{Our code and pre-trained models are publicly available at \url{https://github.com/princeton-nlp/SimCSE}.}


\end{abstract}

\section{Introduction}

\begin{figure*}[!t]
    \centering
    \includegraphics[width=0.98\textwidth]{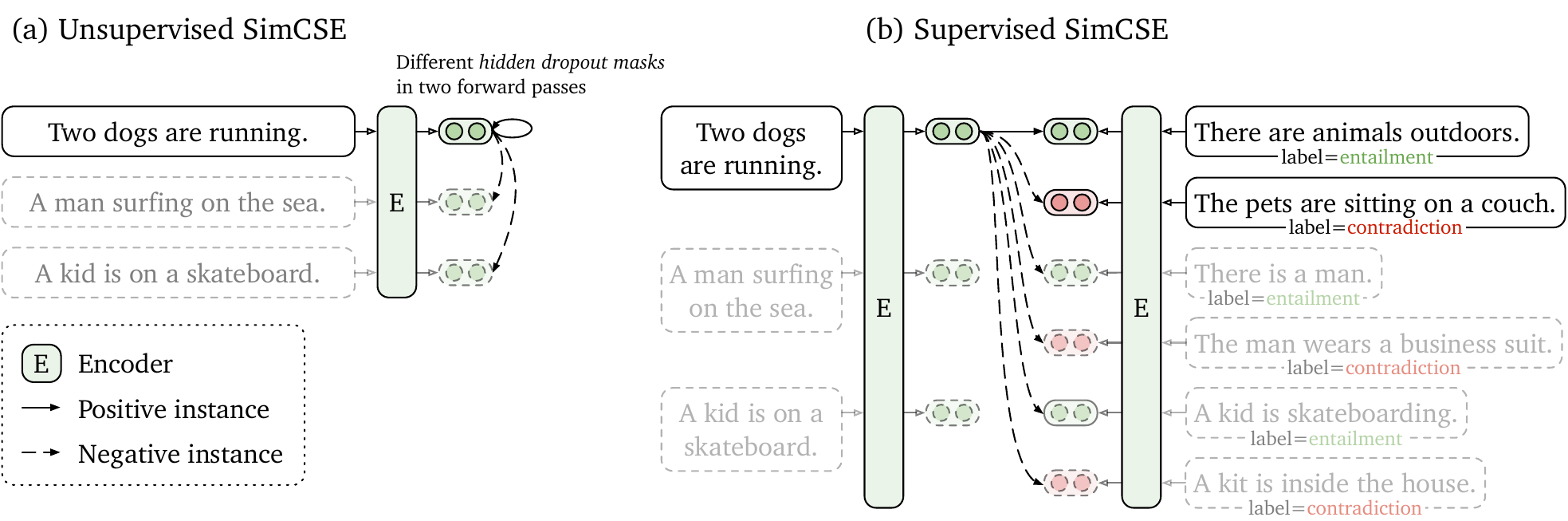}
    \caption{
    (a) Unsupervised \ours predicts the input sentence itself from in-batch negatives, with different hidden dropout masks applied.
    (b) Supervised \ours leverages the NLI datasets and takes the entailment (premise-hypothesis) pairs as positives, and contradiction pairs as well as other in-batch instances as negatives.}
    \label{fig:structure}
\end{figure*}

Learning universal sentence embeddings is a fundamental problem in natural language processing and has been studied extensively in the literature~\cite[][\emph{inter alia}]{kiros2015skip-thought,hill-etal-2016-learning,conneau-etal-2017-supervised-infersent,logeswaran2018an-quick-thought,cer-etal-2018-universal,reimers-gurevych-2019-sentence}. In this work, we advance state-of-the-art sentence embedding methods and demonstrate that a contrastive  objective can be extremely effective when coupled with pre-trained language models such as BERT~\cite{devlin-etal-2019-bert} or RoBERTa~\cite{liu2019roberta}. We present {\ours}, a \underline{sim}ple \underline{c}ontrastive \underline{s}entence \underline{e}mbedding framework,
which can produce superior sentence embeddings, from either unlabeled or labeled data.

Our \ti{unsupervised} {\ours} simply predicts the input sentence itself
with only \ti{dropout}~\cite{srivastava2014dropout} used as noise (Figure~\ref{fig:structure}(a)).
In other words, we pass the same  sentence to the pre-trained encoder \ti{twice}:
by applying the standard dropout twice,
we can obtain two different embeddings as ``positive pairs''.
Then we take other sentences in the same mini-batch as ``negatives'', 
and the model predicts the positive one among the negatives.
Although it may appear strikingly simple, this approach outperforms training objectives such as predicting next sentences~\cite{logeswaran2018an-quick-thought} and
discrete data augmentation (e.g., word deletion and replacement) by a large margin, and even matches previous supervised methods. Through careful analysis, we find that dropout acts as minimal ``data augmentation'' of hidden representations while removing it leads to a representation collapse.

Our \ti{supervised} {\ours} builds upon the recent success of using natural language inference (NLI) datasets for sentence embeddings~\cite{conneau-etal-2017-supervised-infersent,reimers-gurevych-2019-sentence} and incorporates annotated sentence pairs in contrastive learning (Figure~\ref{fig:structure}(b)).
Unlike previous work that casts it as a 3-way classification task (entailment, neutral, and contradiction), we leverage the fact that entailment pairs can be naturally used as positive instances.
We also find that adding corresponding contradiction pairs as hard negatives further improves performance. This simple use of NLI datasets achieves a substantial improvement compared to prior methods using the same datasets.
We also compare to other labeled sentence-pair datasets and find that NLI datasets are especially effective for learning sentence embeddings.

To better understand the strong performance of \ours, we borrow the analysis tool from \citet{wang2020understanding}, which takes \emph{alignment} between semantically-related positive pairs and \emph{uniformity} of the whole representation space to measure the quality of learned embeddings.
Through empirical analysis, we find that our unsupervised SimCSE essentially improves uniformity while avoiding degenerated alignment via dropout noise, thus improving the expressiveness of the representations.
The same analysis shows that the NLI training signal can further improve alignment between positive pairs and produce better sentence embeddings.
We also draw a connection to the recent findings that pre-trained word embeddings suffer from anisotropy~\cite{ethayarajh-2019-contextual,li-etal-2020-sentence}
and prove that---through a spectrum perspective---the contrastive learning objective ``flattens'' the singular value distribution of the sentence embedding space, hence improving uniformity.

We conduct a comprehensive evaluation of \ours
on seven
standard
semantic textual similarity (STS) tasks~\cite{agirre-etal-2012-semeval,agirre-etal-2013-sem,agirre-etal-2014-semeval,agirre-etal-2015-semeval,agirre-etal-2016-semeval,cer-etal-2017-semeval,marelli-etal-2014-sick}
and seven transfer tasks~\cite{conneau-kiela-2018-senteval}.
On the STS tasks,
our unsupervised and supervised models achieve a 76.3\% and 81.6\% averaged Spearman’s correlation respectively using BERT\ba,
a 4.2\% and 2.2\% improvement compared to previous best results.
We also achieve competitive performance on the transfer tasks.
Finally, we identify an incoherent evaluation issue in the literature and consolidate the results of different settings for future work in evaluation of sentence embeddings.



\section{Background: Contrastive Learning}
\label{sec:contrastive_learning}

Contrastive learning aims to learn effective representation by pulling semantically close neighbors together and pushing apart non-neighbors~\cite{hadsell2006dimensionality}.
It assumes a set of paired examples $\mathcal{D} = \{(x_i, x^{+}_i) \}_{i=1}^m$, where $x_i$ and $x_i^+$ are semantically related. We follow the contrastive framework in \citet{chen2020simple} and take a cross-entropy  objective with in-batch negatives~\cite{chen2017sampling,henderson2017efficient}: let $\mf{h}_i$ and $\mf{h}_i^+$ denote the  representations of $x_i$ and $x_i^+$, the training objective for $(x_i, x^{+}_i)$ with a mini-batch of $N$ pairs is:
\begin{equation}
    \label{eq:objective}
    \begin{aligned}
        \ell_i = -\log \frac{e^{\mr{sim}(\mf{h}_i, \mf{h}^+_i)/\tau}}{\sum_{j=1}^N e^{\mr{sim}(\mf{h}_i, \mf{h}_j^+)/\tau}},
    \end{aligned}
\end{equation}
where $\tau$ is a temperature hyperparameter and $\mr{sim}(\mf{h}_1,\mf{h}_2)$ is the cosine similarity $ \frac{\mf{h}_1^\top \mf{h}_2}{\Vert \mf{h}_1\Vert \cdot \Vert \mf{h}_2\Vert}$. In this work, we encode input sentences using a pre-trained language model such as BERT~\cite{devlin-etal-2019-bert} or RoBERTa~\cite{liu2019roberta}: $\mf{h} = f_{\theta}(x)$, and then fine-tune all the parameters using the contrastive learning objective (Eq.~\ref{eq:objective}).


\paragraph{Positive instances.}~~One critical question in contrastive learning is how to construct $(x_i, x_i^+)$ pairs. In visual representations, an effective solution is to take two random transformations of the \ti{same} image (e.g., cropping, flipping, distortion and rotation) as $x_i$ and $x_i^+$~\cite{dosovitskiy2014discriminative}. A similar approach has been recently adopted in language representations~\cite{wu2020clear,meng2021coco} by applying augmentation techniques such as word deletion, reordering, and substitution. However, data augmentation in NLP is inherently difficult because of its discrete nature. As we will see in \S\ref{sec:unsup_simcse},
simply using standard dropout on intermediate representations outperforms these discrete operators.

In NLP, a similar contrastive learning objective has been explored in different contexts~\cite{henderson2017efficient,gillick2019learning,karpukhin-etal-2020-dense}. In these cases, $(x_i, x_i^+)$ are collected from supervised datasets such as question-passage pairs. Because of the distinct nature of $x_i$ and $x_i^+$, these approaches always use a \ti{dual}-encoder framework, i.e., using two independent encoders $f_{\theta_1}$ and $f_{\theta_2}$ for $x_i$ and $x_i^+$. For sentence embeddings, \newcite{logeswaran2018an-quick-thought} also use contrastive learning with a dual-encoder approach, by forming current sentence and next sentence as $(x_i, x^+_i)$.

\paragraph{Alignment and uniformity.} Recently, \citet{wang2020understanding} identify two key properties related to contrastive learning---\emph{alignment} and \emph{uniformity}---and propose to use them to measure the quality of representations.
Given a distribution of positive pairs $p_{\mr{pos}}$, {alignment} calculates expected distance between embeddings of the paired instances (assuming representations are already normalized):
\begin{equation}
    \label{eq:alignment}
    \resizebox{.73\hsize}{!}{%
    $
    \ell_{\mr{align}}\triangleq \underset{(x, x^+)\sim p_{\mr{pos}}}{\mathbb{E}} \Vert f(x) - f(x^+) \Vert^2.
    $
    }
\end{equation}
On the other hand, {uniformity} measures how well the embeddings are uniformly distributed:
\begin{equation}
    \resizebox{.85\hsize}{!}{%
    $
    \label{eq:uniformity}
    \ell_{\mr{uniform}}\triangleq\log \underset{~~~x, y\stackrel{i.i.d.}{\sim} p_{\mr{data}}}{\mathbb{E}}   e^{-2\Vert f(x)-f(y) \Vert^2},
    $
    }
\end{equation}
where $p_{\mr{data}}$ denotes the data distribution. These two metrics are well aligned with the objective of contrastive learning: positive instances should stay close and embeddings for random instances should scatter on the hypersphere. In the following sections, we will also use the two metrics to justify the inner workings of our approaches.



\section{Unsupervised SimCSE}

\label{sec:unsup_simcse}
The idea of unsupervised SimCSE is extremely simple: we take a collection of sentences $\{x_i\}_{i=1}^{m}$ and use $x^+_i = x_i$.
The key ingredient to get this to work with identical positive pairs is through
the use of independently sampled \ti{dropout masks} for $x_i$ and $x^+_i$.
In standard training of Transformers~\cite{vaswani2017attention}, there are dropout masks placed on fully-connected layers as well as attention probabilities (default $p = 0.1$). We denote $\mf{h}_i^z = f_{\theta}(x_i, z)$ where $z$ is a random mask for dropout.
We simply feed the same input to the encoder \ti{twice} and get two embeddings with different dropout masks $z, z'$, and the training objective of SimCSE becomes:
\begin{equation}
    \label{eq:unsup_objective}
    \begin{aligned}
        \ell_i = - \log \frac{e^{\mr{sim}(\mf{h}_i^{z_i}, \mf{h}_i^{z'_i}) / \tau }}{\sum_{j=1}^Ne^{\mr{sim}(\mf{h}_i^{z_i}, \mf{h}_j^{z_j'}) /\tau}},
    \end{aligned}
\end{equation}
for a mini-batch of $N$ sentences. Note that $z$ is just the  \ti{standard} dropout mask in Transformers  and we do not add any additional dropout.


%
%

\begin{table}[t]
    \begin{center}
    \centering
    \small
    \resizebox{0.88\columnwidth}{!}{%
    \begin{tabular}{lccc}
    \toprule
    \tf{Data augmentation} & & & STS-B\\
    \midrule
    None (unsup. SimCSE) & & & \tf{82.5} \\
    \midrule
        Crop & \textit{10\%} & \textit{20\%} & \textit{30\%} \vspace{2pt} \\
        & 77.8 & 71.4  & 63.6\\
    \midrule
       Word deletion & \textit{10\%} & \textit{20\%} & \textit{30\%} \vspace{2pt} \\
          & 75.9 & 72.2 &  68.2 \\
    \midrule
        Delete one word & &&75.9\\
        \tableindent w/o dropout &&&74.2\\
        \multicolumn{3}{l}{Synonym replacement} & 77.4\\
        MLM 15\% & &&62.2\\
    \bottomrule
    \end{tabular}
    }
    \end{center}

    \caption{
        Comparison of data augmentations on STS-B development set (Spearman's correlation).
        \ti{Crop $k$\%}: keep 100-$k$\% of the length;
        \ti{word deletion $k$\%}: delete $k$\% words;
        \ti{Synonym replacement}: use \texttt{nlpaug}~\cite{ma2019nlpaug} to randomly replace one word with its synonym;
        \ti{MLM $k$\%}: use BERT\ba~to replace $k$\% of words.
    }
    \label{tab:data_aug}
\end{table}


%

\begin{table}[t]
    \begin{center}
    \centering
    \small
    \resizebox{0.82\columnwidth}{!}{%
    \begin{tabular}{lcc}
    \toprule
       \tf{Training objective} &  $f_{\theta}$ & $(f_{\theta_1}, f_{\theta_2})$ \\
    \midrule
        Next sentence & 67.1 & 68.9 \\
        Next 3 sentences & 67.4 & 68.8\\
        Delete one word &75.9 & 73.1 \\
        Unsupervised \ours & \tf{82.5} &  80.7\\
    \bottomrule
    \end{tabular}
    }
    \end{center}

    \caption{Comparison of different unsupervised  objectives (STS-B development set, Spearman's correlation). 
    The two columns denote whether we use one encoder or two independent encoders. 
    \ti{Next 3 sentences}: randomly sample one from the next 3 sentences. \ti{Delete one word}: delete one word randomly (see Table~\ref{tab:data_aug}).
    }
    \label{tab:sts_dataset_identical}
\end{table}

\paragraph{Dropout noise as data augmentation.} We view it as a minimal form of data augmentation: the positive pair takes exactly the same sentence, and their embeddings only differ in dropout masks. We compare this approach to other training objectives on the STS-B development set~\cite{cer-etal-2017-semeval}\footnote{
We randomly sample $10^6$ sentences from English Wikipedia and fine-tune BERT\ba~with learning rate = 3e-5, $N$ = 64. In all our experiments, no STS training sets are used.
}. Table~\ref{tab:data_aug} compares our approach to common data augmentation techniques such as crop, word deletion and replacement, which can be viewed as $\mf{h} = f_{\theta}(g(x), z)$ and $g$ is a (random) discrete operator on $x$. We note that even  deleting one word would hurt performance and none of the discrete augmentations outperforms dropout noise.

We also compare this self-prediction training objective to the next-sentence objective used in \newcite{logeswaran2018an-quick-thought}, taking either one encoder or two independent encoders. As shown in Table~\ref{tab:sts_dataset_identical}, we find that {\ours} performs much better than the next-sentence objectives (82.5 vs 67.4 on STS-B) and using one encoder instead of two makes a significant difference in our approach.

\begin{figure}[t]
    \centering
    \includegraphics[width=0.99\linewidth]{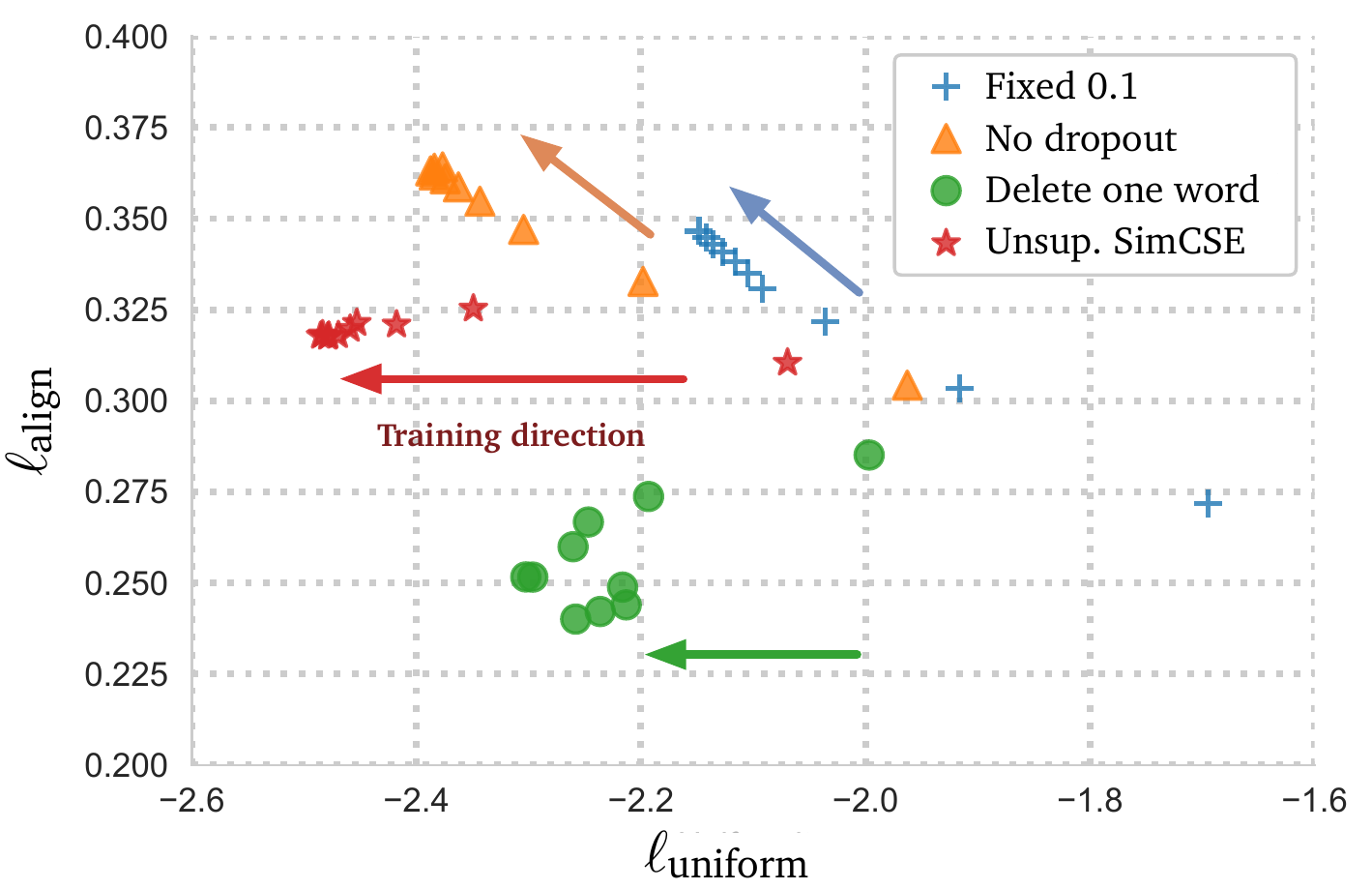}
    \caption{$\ell_{\mr{align}}$-$\ell_{\mr{uniform}}$ plot for unsupervised {\ours}, ``no dropout'',  ``fixed 0.1'',
     and ``delete one word''. We visualize checkpoints every 10 training steps and the arrows indicate the training direction.
    For both $\ell_{\mr{align}}$ and $\ell_{\mr{uniform}}$, \ti{lower numbers are better}.
    }
    \label{fig:identical_align}
    \vspace{-5pt}
\end{figure}

\paragraph{Why does it work?} To further understand the role of dropout noise in unsupervised \ours, we try out different dropout rates in Table~\ref{tab:drop} and observe that all the variants underperform the default dropout probability $p=0.1$ from Transformers.
We find two extreme cases particularly interesting: ``no dropout'' ($p=0$) and ``fixed 0.1'' (using default dropout $p=0.1$ but the same dropout masks for the pair). In both cases, the resulting embeddings for the pair are exactly the same, and it leads to  a dramatic performance degradation.
We take the checkpoints of these models every 10 steps during training and visualize the alignment and uniformity metrics\footnote{We take STS-B pairs with a score higher than 4 as $p_{\mr{pos}}$ and all STS-B sentences as  $p_{\mr{data}}$.} in Figure~\ref{fig:identical_align}, along with a simple data augmentation model ``delete one word''.
As clearly shown, starting from pre-trained checkpoints, all  models greatly improve uniformity. However, the alignment of the two special variants also degrades drastically, while our unsupervised \ours keeps a steady alignment, thanks to the use of dropout noise.
It also demonstrates that starting from a pre-trained checkpoint is crucial, for it provides good initial alignment.
At last, ``delete one word'' improves the alignment yet achieves a smaller gain on the uniformity metric, and eventually underperforms unsupervised \ours.


%
%

\begin{table}[t]
    \begin{center}
    \centering
    \small
    \resizebox{0.8\columnwidth}{!}{%
    \begin{tabular}{lcccc}
    \toprule
       {$p$} & \emph{0.0} & \emph{0.01} & \emph{0.05} & \emph{0.1}\vspace{2pt}  \\
       STS-B & 71.1 &72.6 &  81.1  & \tf{82.5}\\
       \midrule
       {$p$} & \emph{0.15} & \emph{0.2} & \emph{0.5} & \emph{Fixed 0.1}\vspace{2pt}\\
       STS-B & 81.4 &80.5 & 71.0 & 43.6 \\
    \bottomrule
    \end{tabular}
    }
    \end{center}

    \caption{
        Effects of different dropout probabilities $p$ on the STS-B development set (Spearman's correlation, BERT\ba).  \textit{Fixed 0.1}: default 0.1 dropout rate but apply the same dropout mask on both $x_i$ and $x^+_i$.
    }
    \label{tab:drop}
    \vspace{-4pt}
\end{table}

\section{Supervised SimCSE}

\label{sec:sup_simcse}

We have demonstrated that adding dropout noise is able to keep a good alignment for positive pairs $(x, x^+) \sim p_{\text{pos}}$. In this section, we study whether we can leverage supervised datasets to provide better training signals for improving alignment of our approach. Prior work~\cite{conneau-etal-2017-supervised-infersent,reimers-gurevych-2019-sentence} has demonstrated that supervised natural language inference (NLI) datasets~\cite{bowman-etal-2015-large-snli,williams-etal-2018-broad-mnli} are effective for learning sentence embeddings, by predicting  whether the relationship between two sentences is \ti{entailment}, \ti{neutral} or \ti{contradiction}. In our contrastive learning framework, we instead directly take $(x_i, x_i^+)$ pairs from supervised datasets and use them to optimize Eq.~\ref{eq:objective}.

\paragraph{Choices of labeled data.} We first explore which supervised datasets are especially suitable for constructing positive pairs $(x_i, x^+_i)$. We experiment with a number of datasets with sentence-pair
examples,
including 1) QQP\footnote{\url{https://www.quora.com/q/quoradata/}}:
Quora question pairs; 2) Flickr30k~\cite{young-etal-2014-image-flickr30k}: each image is annotated with 5 human-written captions and we consider any two captions of the same image as a positive pair; 3) ParaNMT~\cite{wieting-gimpel-2018-paranmt}: a large-scale back-translation paraphrase dataset\footnote{ParaNMT is automatically constructed by machine translation systems. Strictly speaking, we should not call it ``supervised''. It underperforms our unsupervised {\ours} though.}; and finally 4) NLI datasets: SNLI~\cite{bowman-etal-2015-large-snli} and MNLI~\cite{williams-etal-2018-broad-mnli}.

We train the contrastive learning model (Eq.~\ref{eq:objective}) with different datasets and compare the results in Table~\ref{tab:sts_dataset}. For a fair comparison, we also run experiments with the same \# of training pairs.
Among all the options, using entailment pairs from the NLI (SNLI + MNLI) datasets performs the best. We think this is reasonable, as the NLI datasets consist of high-quality and crowd-sourced pairs. Also, human annotators are expected to write the hypotheses manually based on the premises and two sentences tend to have less lexical overlap. For instance, we find that the lexical overlap (F1 measured between two bags of words) for the entailment pairs (SNLI + MNLI) is 39\%, while they are 60\% and 55\% for QQP and ParaNMT.

\paragraph{Contradiction as hard negatives.}
Finally, we further take the advantage of the NLI datasets by using its contradiction pairs as hard negatives\footnote{We also experimented with adding neutral hypotheses as hard negatives. See Section~\ref{sec:hard-negatives} for more discussion.}.
In NLI datasets, given one premise, annotators are required to manually write one sentence that is absolutely true (\emph{entailment}), one that might be true (\emph{neutral}), and one that is definitely false (\emph{contradiction}). Therefore, for each premise and its entailment hypothesis, there is an accompanying contradiction hypothesis\footnote{In fact, one premise can have multiple contradiction hypotheses. In our implementation, we only sample one as the hard negative and we did not find a difference by using more.} (see Figure~\ref{fig:structure} for an example).

Formally, we extend $(x_i, x_i^+)$ to $(x_i, x_i^+, x_i^-)$, where $x_i$ is the premise, $x_i^+$ and $x_i^-$ are entailment and contradiction hypotheses. The training objective $\ell_i$ is then defined by ($N$ is mini-batch size):
    \vspace{-8pt}
\begin{equation}
    \label{eq:sup_objective}
    \begin{aligned}
        - \log \frac{e^{\mr{sim}(\mf{h}_i,\mf{h}_i^+ )/ \tau }}{\sum_{j=1}^N\left(e^{\mr{sim}(\mf{h}_i,\mf{h}_j^+)/\tau}+e^{\mr{sim}(\mf{h}_i,\mf{h}_j^-)/ \tau}\right)}.
    \end{aligned}
\end{equation}
As shown in Table~\ref{tab:sts_dataset}, adding hard negatives can further improve performance (84.9 $\rightarrow$ 86.2) and this is our final supervised {\ours}. We also tried to add the ANLI dataset~\cite{nie-etal-2020-adversarial} or combine it with our unsupervised {\ours} approach, but didn't find a meaningful improvement. We also considered a dual encoder framework in supervised {\ours} and it hurt performance (86.2 $\rightarrow$ 84.2).


\begin{table}[t]
    \begin{center}
    \centering
    \small
    \resizebox{0.8\columnwidth}{!}{%
    \begin{tabular}{lcc}
    \toprule
       \tf{Dataset} & sample  & {full} \\
    \midrule
        Unsup. {\ours} (1m) & - & 82.5 \\
        \midrule
        QQP (134k) & 81.8 & 81.8 \\
        Flickr30k (318k) & 81.5 & 81.4  \\
        ParaNMT (5m) & 79.7 & 78.7 \\
        SNLI+MNLI \\
        \tableindent entailment (314k) & \tf{84.1} &  \tf{84.9} \\
        \tableindent neutral (314k)\footnote{placeholder} &82.6 &82.9 \\
        \tableindent contradiction (314k) & 77.5 & 77.6 \\
        \tableindent all (942k) & 81.7 & 81.9 \\
        \midrule
        SNLI+MNLI \\
        \tableindent entailment + hard neg. & - & \tf{86.2} \\
        \tableindent + ANLI (52k) &  - & 85.0 \\
    \bottomrule
    \end{tabular}
    }
    \end{center}

    \caption{
        Comparisons of different supervised datasets as positive pairs. Results are Spearman's correlations on the STS-B development set using BERT\ba~(we use the same hyperparameters as the final SimCSE model). Numbers in brackets denote the \# of pairs. \ti{Sample}: subsampling 134k positive pairs for a fair comparison among datasets; \ti{full}: using the full dataset. In the last block, we use entailment pairs as positives and contradiction pairs as hard negatives (our final model).
    }
    \label{tab:sts_dataset}
\end{table}


\section{Connection to Anisotropy}
\label{sec:claim}

Recent work identifies an \ti{anisotropy} problem in language representations~\cite{ethayarajh-2019-contextual,li-etal-2020-sentence}, i.e., the learned embeddings occupy a narrow cone in the vector space, which severely limits their expressiveness. \citet{gao2018representation} demonstrate that language models trained with tied input/output embeddings lead to anisotropic word embeddings, and this is further observed by \citet{ethayarajh-2019-contextual} in pre-trained contextual representations. \citet{Wang2020Improving} show that singular values of the word embedding matrix in a language model decay drastically: except for a few dominating singular values, all others are close to zero.

\footnotetext[8]{Though our final model only takes entailment pairs as positive instances, here we also try taking neutral and contradiction pairs from the NLI datasets as positive pairs.}

A simple way to alleviate the problem is post-processing, either to eliminate the dominant principal components~\cite{arora2017simple,mu2018allbutthetop}, or to map embeddings to an isotropic distribution~\cite{li-etal-2020-sentence,su2021whitening}.
Another common solution is to add regularization during training~\cite{gao2018representation,Wang2020Improving}.
In this work, we show that---both theoretically and empirically---the contrastive objective can also alleviate the anisotropy problem.

The anisotropy problem is naturally connected to \emph{uniformity} \cite{wang2020understanding},
both highlighting that embeddings should be evenly distributed in the space.
Intuitively, optimizing the contrastive learning objective can improve uniformity (or ease the anisotropy problem), as the objective pushes negative instances apart.
Here, we take a singular spectrum perspective---which is a common practice in analyzing word embeddings~\cite{mu2018allbutthetop,gao2018representation,Wang2020Improving}, and show
that the contrastive objective can ``flatten'' the singular value distribution of sentence embeddings  and make the representations more isotropic.


Following \citet{wang2020understanding}, the asymptotics of the contrastive learning objective (Eq.~\ref{eq:objective}) can be expressed by the following equation when the number of negative instances approaches infinity (assuming $f(x)$ is normalized):
\begin{equation}
\label{eq:asymptotic}
\resizebox{.85\hsize}{!}{%
$
\begin{aligned}
    &-\frac{1}{\tau}\underset{(x,x^+)\sim p_{\mr{pos}}}{\mathbb{E}}\left[f(x)^\top f(x^+)\right] \\
    &+\underset{x\sim p_{\mr{data}}}{\mathbb{E}}\left[\log\underset{x^-\sim p_{\mr{data}}}{\mathbb{E}}\left[e^{f(x)^\top f(x^-)/\tau}\right]\right],
\end{aligned}
$}
\end{equation}
where the first term keeps positive instances similar and the second pushes negative pairs apart. When $p_{\mr{data}}$ is uniform over finite samples $\{x_i\}_{i=1}^m$, with $\mf{h}_i = f(x_i)$,  we can derive the following formula from the second term with Jensen's inequality:
\begin{equation}
\label{eq:jensen}
\resizebox{.8\hsize}{!}{%
$
\begin{aligned}
&\underset{x\sim p_{\mr{data}}}{\mathbb{E}}\left[\log\underset{x^-\sim p_{\mr{data}}}{\mathbb{E}}\left[e^{f(x)^\top f(x^-)/\tau}\right]\right]\\
=&\frac{1}{m}\sum_{i=1}^m\log\left(\frac{1}{m}\sum_{j=1}^me^{\mf{h}_i^\top \mf{h}_j/\tau}\right) \\
\geq & \frac{1}{\tau m^2}\sum_{i=1}^m\sum_{j=1}^m \mf{h}_i^\top \mf{h}_j.
\end{aligned}
$}
\end{equation}
Let $\mf{W}$ be the sentence embedding matrix corresponding to $\{x_i\}_{i=1}^m$, i.e., the $i$-th row of $\mf{W}$ is $\mf{h}_i$. Optimizing the second term in Eq.~\ref{eq:asymptotic} essentially minimizes an upper bound of the summation of all elements in $\mf{W}\mf{W}^\top$, i.e.,
$\mr{Sum}(\mf{W}\mf{W}^\top) = \sum_{i=1}^m \sum_{j=1}^m\mf{h}_i^\top\mf{h}_j$.

Since we normalize $\mf{h}_i$, all elements on the diagonal of $\mf{W}\mf{W}^\top$ are $1$ and then $\mr{tr}(\mf{W}\mf{W}^\top)$ (the sum of all eigenvalues) is a constant.
According to \citet{MERIKOSKI1984177}, if all elements in $\mf{W}\mf{W}^\top$ are positive,
which is the case in most times according to Figure~\ref{fig:histogram},
then $\mr{Sum}(\mf{W}\mf{W}^\top)$ is an upper bound for the largest eigenvalue of $\mf{W} \mf{W}^\top$.
When minimizing the second term in Eq.~\ref{eq:asymptotic}, we reduce the top eigenvalue of $\mf{W} \mf{W}^\top$ and inherently ``flatten'' the singular spectrum of the embedding space. Therefore, contrastive learning is expected to alleviate the representation degeneration problem and improve uniformity of sentence embeddings. 

Compared to post-processing methods in \citet{li-etal-2020-sentence,su2021whitening}, which only aim to encourage isotropic representations, contrastive learning also optimizes for aligning positive pairs by the first term in Eq.~\ref{eq:asymptotic}, which is the key to the success of \ours. A quantitative analysis is given in \S\ref{sec:analysis}.


\section{Experiment}
\label{sec:experiment}


\begin{table*}[t]
    \begin{center}
    \centering
    \small
    \begin{tabular}{lcccccccc}
    \toprule
       \tf{Model} & \tf{STS12} & \tf{STS13} & \tf{STS14} & \tf{STS15} & \tf{STS16} & \tf{STS-B} & \tf{SICK-R} & \tf{Avg.} \\
    \midrule
    \midrule
        \multicolumn{9}{c}{\it{Unsupervised models}}\\
    \midrule
        GloVe embeddings (avg.)$^\clubsuit$ & 55.14 & 70.66 & 59.73 & 68.25 & 63.66 & 58.02 & 53.76 & 61.32 \\
        BERT\ba~(first-last avg.) & 39.70&	59.38&	49.67&	66.03&	66.19&	53.87&	62.06&	56.70\\ 
        BERT\ba-flow & 58.40&	67.10&	60.85&	75.16&	71.22&	68.66&	64.47&	66.55 \\ 
        BERT\ba-whitening & 57.83& 66.90 & 60.90 & 75.08& 71.31& 68.24& 63.73& 66.28\\ 
        IS-BERT\ba$^\heartsuit$ & 56.77 & 69.24 & 61.21 & 75.23 & 70.16 & 69.21 & 64.25 & 66.58 \\
        CT-BERT\ba & 61.63 & 76.80 & 68.47 & 77.50 & 76.48 & 74.31 & 69.19 &72.05 \\
        $*$ \ours-BERT\ba & \tf{68.40}&	\tf{82.41} &	\tf{74.38}&	\tf{80.91}&	\tf{78.56}&	\tf{76.85}&	\tf{72.23}&	\tf{76.25} \\
        \midrule
        RoBERTa\ba~(first-last avg.) & 40.88&	58.74&	49.07&	65.63&	61.48&	58.55&	61.63&	56.57\\
        RoBERTa\ba-whitening & 46.99& 63.24&	57.23&	71.36&	68.99&	61.36&	62.91& 61.73\\ 
        DeCLUTR-RoBERTa\ba & 52.41 & 75.19& 65.52 & 77.12 & 78.63 & 72.41 & \tf{68.62} & 69.99\\
        $*$ \ours-RoBERTa\ba & \tf{70.16} &	\tf{81.77}	&	\tf{73.24}	&	\tf{81.36}	&	\tf{80.65}	&	\tf{80.22}	&	68.56 & \tf{76.57}\\
        $*$ \ours-RoBERTa\la & \tf{72.86}	& \tf{83.99}	& \tf{75.62}	& \tf{84.77}	& \tf{81.80}	& \tf{81.98}	& \tf{71.26} & \tf{78.90}\\
    \midrule
    \midrule
        \multicolumn{9}{c}{\it{Supervised models}}\\
    \midrule
        InferSent-GloVe$^\clubsuit$ & 52.86 & 66.75 & 62.15 & 72.77 & 66.87 & 68.03 & 65.65 & 65.01\\
        Universal Sentence Encoder$^\clubsuit$ & 64.49 & 67.80 & 64.61 & 76.83 & 73.18 & 74.92 & 76.69 & 71.22 \\
        SBERT\ba$^\clubsuit$ & 70.97 & 76.53 & 73.19 & 79.09 & 74.30 & 77.03 & 72.91 & 74.89\\
        SBERT\ba-flow  & 69.78&	77.27&	74.35&	82.01&	77.46&	79.12&	76.21&	76.60 \\ 
        SBERT\ba-whitening & 69.65&	77.57&	74.66&	82.27&	78.39&	79.52&	76.91&	77.00 \\
        CT-SBERT\ba & 74.84	&83.20 &78.07 &83.84 &77.93 &81.46 &76.42 & 79.39\\
        $*$ \ours-BERT\ba & \tf{75.30}&	\tf{84.67}&	\tf{80.19} &	\tf{85.40} &	\tf{80.82} &	\tf{84.25} &	\tf{80.39} &	\tf{81.57} \\
        \midrule
        SRoBERTa\ba$^\clubsuit$ & 71.54 & 72.49 & 70.80 & 78.74 & 73.69 & 77.77 & 74.46 & 74.21\\
        SRoBERTa\ba-whitening & 70.46&	77.07&	74.46&	81.64&	76.43&	79.49&	76.65&	76.60 \\
                $*$ \ours-RoBERTa\ba  & \tf{76.53}&	\tf{85.21}&	\tf{80.95}&	\tf{86.03}&	\tf{82.57}&	\tf{85.83}&	\tf{80.50}&	\tf{82.52}\\ 
        $*$ \ours-RoBERTa\la & \tf{77.46}&	\tf{87.27}&	\tf{82.36}&	\tf{86.66}&	\tf{83.93}&	\tf{86.70}&	\tf{81.95}&	\tf{83.76}\\
    \bottomrule
    \end{tabular}
    \end{center}

    \caption{
        Sentence embedding performance on STS tasks (Spearman's correlation, ``all'' setting).
        We highlight the highest numbers among models with the same pre-trained encoder.
        $\clubsuit$: results from \citet{reimers-gurevych-2019-sentence};
        $\heartsuit$: results from \citet{zhang-etal-2020-unsupervised};
        all other results are reproduced or reevaluated by ourselves.
        For BERT-flow~\cite{li-etal-2020-sentence} and whitening~\cite{su2021whitening}, we only report the ``NLI'' setting (see Table~\ref{tab:target_vs_nli}).
    }
    \label{tab:main_sts}
    \vspace{-5pt}
\end{table*}

\subsection{Evaluation Setup}
\label{sec:evaluation}

We conduct our experiments on 7 semantic textual similarity (STS) tasks.
Note that all our STS experiments are fully \tf{unsupervised} and no STS training sets are used.
Even for supervised SimCSE, we simply mean that we take extra labeled datasets for training, following previous work~\cite{conneau-etal-2017-supervised-infersent}.
We also evaluate 7 transfer learning tasks and provide detailed results in Appendix~\ref{sec:transfer}. We share a similar sentiment with \newcite{reimers-gurevych-2019-sentence} that the main goal of sentence embeddings is to cluster semantically similar sentences and hence take STS as the main result.

\paragraph{Semantic textual similarity tasks.}
We evaluate on 7 STS tasks: STS 2012--2016~\cite{agirre-etal-2012-semeval,agirre-etal-2013-sem,agirre-etal-2014-semeval,agirre-etal-2015-semeval,agirre-etal-2016-semeval},
STS Benchmark~\cite{cer-etal-2017-semeval} and
SICK-Relatedness~\cite{marelli-etal-2014-sick}.
When comparing to previous work, we identify invalid comparison patterns in published papers in the evaluation settings, including (a) whether to use an additional regressor, (b) Spearman's vs Pearson's correlation, and (c) how the results are aggregated (Table~\ref{tab:metric_used}). We discuss the detailed differences in Appendix~\ref{sec:sts_discrepancy} and choose to follow the setting of \citet{reimers-gurevych-2019-sentence} in our evaluation (no additional regressor, Spearman's correlation, and ``all'' aggregation). We also report our replicated study of previous work as well as our results evaluated in a different setting in Table~\ref{tab:replicated_results} and Table~\ref{tab:wmean_sts}. We call for unifying the setting in evaluating sentence embeddings for future research.

\paragraph{Training details.} We start from pre-trained checkpoints of BERT \cite{devlin-etal-2019-bert} (uncased) or RoBERTa \cite{liu2019roberta} (cased) and take the \cls~representation as the sentence embedding\footnote{There is an MLP layer over \cls~in BERT's original implementation and we keep it with random initialization.} (see \S\ref{sec:ablation} for comparison between different pooling methods).
We train unsupervised SimCSE on $10^6$ randomly sampled sentences from English Wikipedia, and train supervised SimCSE on the combination of MNLI and SNLI datasets (314k).
More training details can be found in Appendix~\ref{sec:training_details}.

\subsection{Main Results}
We compare unsupervised and supervised
\ours to previous state-of-the-art sentence embedding methods on STS tasks.
Unsupervised baselines include
average GloVe embeddings~\cite{pennington-etal-2014-glove}, average BERT or RoBERTa embeddings\footnote{
    Following~\citet{su2021whitening}, we take the average of the first and the last layers, which is better than only taking the last.
}, and post-processing methods such as BERT-flow~\cite{li-etal-2020-sentence} and BERT-whitening~\cite{su2021whitening}.
We also compare to several recent methods using a contrastive objective, including
1) IS-BERT~\cite{zhang-etal-2020-unsupervised}, which maximizes the agreement between global and local features; 2) DeCLUTR~\cite{giorgi2020declutr}, which takes different spans from the same document as positive pairs; 3) CT~\cite{carlsson2021semantic}, which aligns embeddings of the same sentence from two different encoders.\footnote{
    We do not compare to CLEAR~\cite{wu2020clear}, because they use their own version of pre-trained models, and the numbers appear to be much lower. Also note that CT is a concurrent work to ours.
}
Other supervised methods include InferSent~\cite{conneau-etal-2017-supervised-infersent}, Universal Sentence Encoder~\cite{cer-etal-2018-universal}, and SBERT/SRoBERTa~\cite{reimers-gurevych-2019-sentence} with post-processing methods (BERT-flow, whitening, and CT).
We provide more details of these baselines in Appendix~\ref{sec:baseline}.

Table~\ref{tab:main_sts} shows the evaluation results on 7 STS tasks.
\ours can substantially improve results on all the datasets with or without extra NLI supervision, greatly outperforming the previous state-of-the-art models.
Specifically, our unsupervised \ours-BERT\ba~improves the previous best averaged Spearman's correlation from 72.05\% to 76.25\%, even comparable to supervised baselines.
When using NLI datasets, \ours-BERT\ba~further pushes the state-of-the-art results to 81.57\%. The gains are more pronounced on RoBERTa encoders, and our supervised SimCSE achieves 83.76\% with RoBERTa\la.

In Appendix~\ref{sec:transfer}, we show that SimCSE also achieves on par or better transfer task performance compared to existing work, and an auxiliary MLM objective can further boost performance.

\subsection{Ablation Studies}
\label{sec:ablation}

We investigate the impact of different pooling methods and hard negatives. All reported results in this section are based on the STS-B development set. We provide more ablation studies (normalization, temperature, and MLM objectives) in Appendix~\ref{sec:app_ablation}.


\begin{table}[t]
    \begin{center}
    \centering
    \small
    \resizebox{0.8\columnwidth}{!}{%
    \begin{tabular}{lcc}
    \toprule
        \tf{Pooler} & \tf{Unsup.} & \tf{Sup.} \\
    \midrule
        \cls \\
        \tableindent \tableindent w/ MLP & 81.7 & \tf{86.2}\\
        \tableindent \tableindent w/ MLP (train) & \tf{82.5} & 85.8\\
        \tableindent \tableindent w/o MLP& 80.9 & \tf{86.2}\\
        First-last avg. &  81.2 & 86.1\\
    \bottomrule
    \end{tabular}
    }
    \end{center}

    \caption{
        Ablation studies of different pooling methods in unsupervised and supervised SimCSE. \emph{\cls~w/ MLP (train)}: using MLP on \cls~during training but removing it during testing. The results are based on the development set of STS-B using BERT\ba.
    }
    \label{tab:pooler}
\end{table}

\paragraph{Pooling methods.} \citet{reimers-gurevych-2019-sentence,li-etal-2020-sentence} show that taking the average embeddings of pre-trained models (especially from both the first and last layers) leads to better performance than \cls.
Table~\ref{tab:pooler} shows the comparison between different pooling methods in both unsupervised and supervised SimCSE. For \cls~representation, the original BERT implementation takes an extra MLP layer on top of it. Here, we consider three different settings for \cls: 1) keeping the MLP layer; 2) no MLP layer; 3) keeping MLP during training but removing it at testing time. We find that for unsupervised SimCSE, taking \cls~representation with MLP only during training works the best; for supervised SimCSE,
different pooling methods do not matter much.
By default, we take \cls with MLP (train) for unsupervised SimCSE and \cls with MLP for supervised SimCSE.

%


\begin{table}[t]
    \begin{center}
    \centering
    \resizebox{1.0\columnwidth}{!}{%
    \begin{tabular}{lccccc}
    \toprule
        \multirow{2}{*}{\tf{Hard neg}} &\multirow{2}{*}{ N/A} & \multicolumn{3}{c}{\multirow{2}{*}{Contradiction}} & Contra.+\\
        &&&&& Neutral \\
        $\alpha$ & - & 0.5 & 1.0 & 2.0 & 1.0 \\
    \midrule
        \tf{STS-B} &  84.9 & 86.1 & \tf{86.2} & \tf{86.2} & 85.3 \\
    \bottomrule
    \end{tabular}
    }
    \end{center}

    \caption{
        STS-B development results with different hard negative policies. ``N/A'': no hard negative.
    }
    \label{tab:hard_neg}
\end{table}

\paragraph{Hard negatives.}
\label{sec:hard-negatives}
Intuitively, it may be beneficial to differentiate hard negatives (contradiction examples) from other in-batch negatives. Therefore, we extend our training objective defined in Eq.~\ref{eq:sup_objective} to incorporate weighting of different negatives:
\begin{equation}
    \label{eq:weighted_sup_objective}
\resizebox{.89\hsize}{!}{%
$
    \begin{aligned}
        - \log \frac{e^{\mr{sim}(\mf{h}_i,\mf{h}_i^+ )/ \tau }}{\sum_{j=1}^N\left(e^{\mr{sim}(\mf{h}_i,\mf{h}_j^+)/\tau}+\alpha^{\mathbbm{1}^j_i}e^{\mr{sim}(\mf{h}_i,\mf{h}_j^-)/ \tau}\right)},
    \end{aligned}
$
}
\end{equation}
where $\mathbbm{1}^j_i\in\{0,1\}$ is an indicator that equals  1 if and only if $i = j$. We train \ours with different values of $\alpha$ and evaluate the trained models on the development set of STS-B. We also consider taking neutral hypotheses as hard negatives. As shown in Table~\ref{tab:hard_neg}, $\alpha = 1$ performs the best, and neutral hypotheses do not bring further gains.

\section{Analysis}
\label{sec:analysis}
In this section, we conduct further analyses to understand the inner workings of {\ours}.

\begin{figure}[t]
    \centering
    \includegraphics[width=0.99\linewidth]{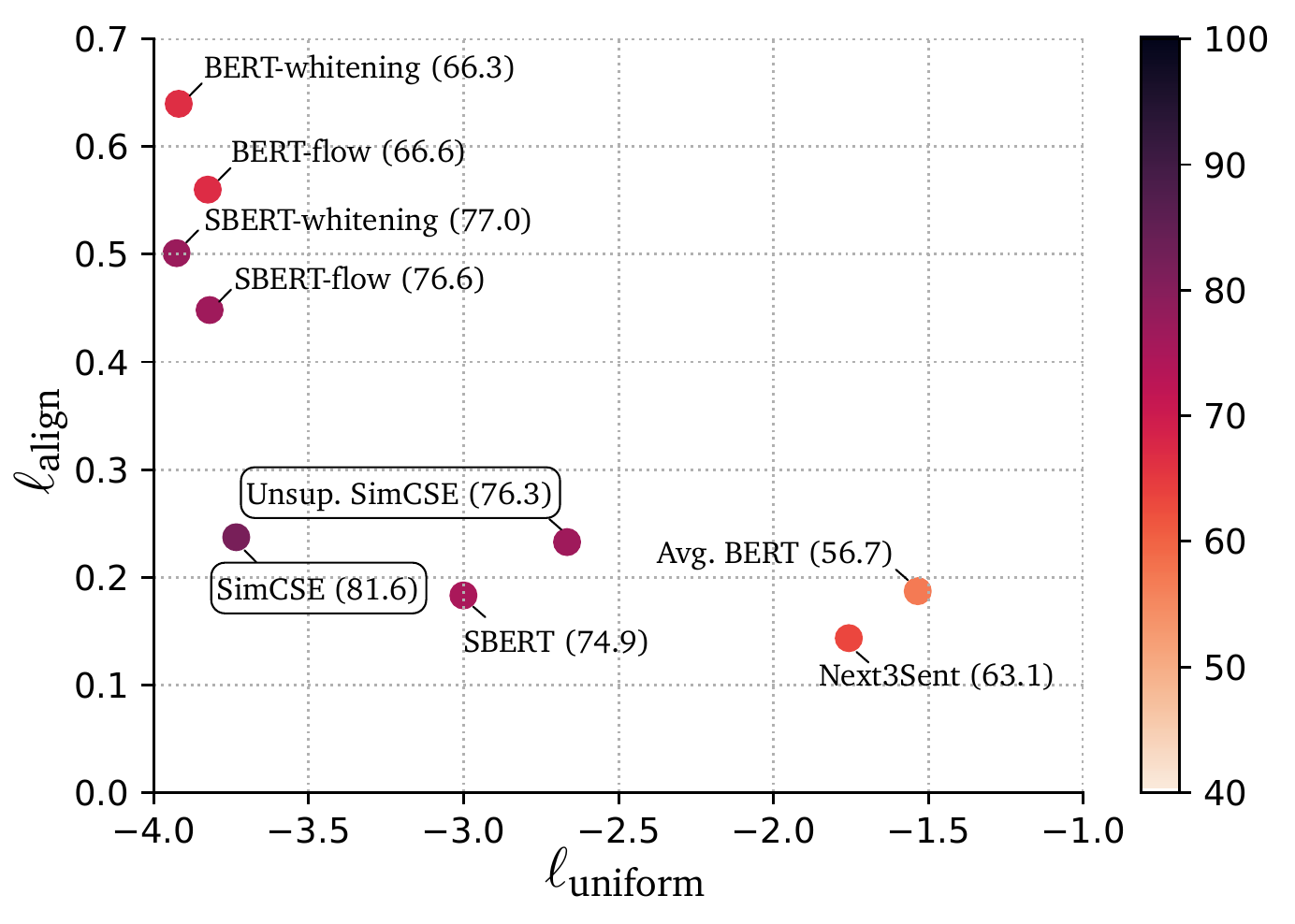}
    \caption{$\ell_{\mr{align}}$-$\ell_{\mr{uniform}}$ plot of models based on BERT\ba. Color of points and numbers in brackets represent average STS performance (Spearman's correlation). \ti{Next3Sent}: ``next 3 sentences'' from Table~\ref{tab:sts_dataset_identical}.}
    \label{fig:align}
    \vspace{-5pt}
\end{figure}

\paragraph{Uniformity and alignment.}
Figure~\ref{fig:align} shows uniformity and alignment of different sentence embedding models along with their averaged STS results.
In general, models which have both better alignment and uniformity achieve better performance, confirming the findings in~\citet{wang2020understanding}.
We also observe that (1) though pre-trained embeddings have good alignment, their uniformity is poor (i.e., the embeddings are highly anisotropic); (2) post-processing methods like BERT-flow and BERT-whitening greatly improve uniformity but also suffer a degeneration in alignment; (3) unsupervised \ours effectively improves uniformity of pre-trained embeddings whereas keeping a good alignment; (4) incorporating supervised data in \ours further amends alignment.
In Appendix~\ref{sec:singular_value}, we further show that \ours can effectively flatten singular value distribution of pre-trained embeddings.
In Appendix~\ref{sec:cos-sim-dis}, we demonstrate that SimCSE provides more distinguishable cosine similarities between different sentence pairs.

%
%


\paragraph{Qualitative comparison.} We conduct a small-scale retrieval experiment using SBERT\ba~and \ours-BERT\ba. We use 150k captions from Flickr30k dataset and take any random sentence as query to retrieve similar sentences (based on cosine similarity). As several examples shown in Table~\ref{tab:retrieval}, the retrieved sentences by {\ours} have a higher quality compared to those retrieved by SBERT.

\begin{table*}[t]
    \centering
    \resizebox{0.98\textwidth}{!}{%
    \begin{tabular}{l|l|l}
        \toprule
        & \tf{SBERT\ba} & \tf{Supervised \ours-BERT\ba}  \\
        \midrule
        \multicolumn{3}{l}{\tf{Query}: A man riding a small boat in a harbor.} \\
        \midrule
        \#1 & A group of men traveling over the ocean in a small boat. & A man on a moored blue and white boat. \\
        \#2 & Two men sit on the bow of a colorful boat. & A man is riding in a boat on the water. \\
        \#3 & A man wearing a life jacket is in a small boat on a lake. & A man in a blue boat on the water.\\
        \midrule
        \multicolumn{3}{l}{\tf{Query}: A dog runs on the green grass near a wooden fence.} \\
        \midrule
        \#1& A dog runs on the green grass near a grove of trees.&The dog by the fence is running on the grass.\\
        \#2&  A brown and white dog runs through the green grass.& Dog running through grass in fenced area.\\
        \#3& The dogs run in the green field. & A dog runs on the green grass near a grove of trees.\\
        \bottomrule
    \end{tabular}
    }
    \caption{Retrieved top-3 examples by SBERT and supervised \ours from Flickr30k (150k sentences). }
    \label{tab:retrieval}
    \vspace{-5pt}
\end{table*}


\section{Related Work}
\label{sec:related}

Early work in sentence embeddings builds upon the distributional hypothesis by predicting surrounding sentences of a given one~\cite{kiros2015skip-thought,hill-etal-2016-learning,logeswaran2018an-quick-thought}.
\citet{pagliardini-etal-2018-unsupervised} show that simply augmenting the idea of word2vec~\cite{Mikolov2013DistributedRO} with n-gram embeddings leads to strong results.
Several recent (and concurrent) approaches adopt contrastive objectives \cite{zhang-etal-2020-unsupervised,giorgi2020declutr,wu2020clear,meng2021coco,carlsson2021semantic,kim2021self,yan2021consert}
by taking different views---from data augmentation or different copies of models---of the same sentence or document.
Compared to these work, \ours uses the simplest idea by taking different outputs of the same sentence from standard dropout, and performs the best on STS tasks.

Supervised sentence embeddings are promised to have stronger performance compared to unsupervised counterparts. \citet{conneau-etal-2017-supervised-infersent} propose to fine-tune a Siamese model on NLI datasets, which is further extended to other encoders or pre-trained models~\cite{cer-etal-2018-universal,reimers-gurevych-2019-sentence}.
Furthermore, \citet{wieting-gimpel-2018-paranmt,wieting-etal-2020-bilingual} demonstrate that bilingual and back-translation corpora provide useful supervision for learning semantic similarity.
Another line of work focuses on regularizing embeddings~\cite{li-etal-2020-sentence,su2021whitening,huang2021whiteningbert}
to alleviate the representation degeneration problem (as discussed in \S\ref{sec:claim}), and yields substantial improvement over pre-trained language models.


\section{Conclusion}
In this work, we propose \ours, a simple contrastive learning framework, which
greatly improves state-of-the-art sentence embeddings on semantic textual similarity tasks. We present an unsupervised approach which predicts input sentence itself with dropout noise  and a supervised approach utilizing NLI datasets.
We further justify the inner workings of our approach by analyzing alignment and uniformity of \ours~along with other baseline models. We believe that our contrastive objective, especially the unsupervised one, may have a broader application in NLP. It provides a new perspective on data augmentation with text input, and can be extended to other continuous representations and integrated in language model pre-training.

\section*{Acknowledgements}

We thank Tao Lei, Jason Lee, Zhengyan Zhang, Jinhyuk Lee, Alexander Wettig, Zexuan Zhong, and the members of the Princeton NLP group for helpful discussion and valuable feedback. This research is supported by a Graduate Fellowship at Princeton University and a gift award from Apple.


%


\newpage
\bibliography{custom}
\bibliographystyle{acl_natbib}

\clearpage

\appendix

\counterwithin{figure}{section}
\counterwithin{table}{section}


%
%
%
%

\section{Training Details}
\label{sec:training_details}
We implement \ours with \texttt{transformers} package~\cite{wolf-etal-2020-transformers}.
For supervised \ours, we train our models for $3$ epochs, evaluate the model every $250$ training steps on the development set of STS-B and keep the best checkpoint for the final evaluation on test sets.
We do the same for the unsupervised \ours, except that we train the model for one epoch.
We carry out grid-search of batch size $\in \{64, 128, 256, 512\}$ and learning rate $\in \{$1e-5, 3e-5, 5e-5$\}$ on STS-B development set and adopt the hyperparameter settings in Table~\ref{tab:hyper}. We find that SimCSE is not sensitive to batch sizes as long as tuning the learning rates accordingly, which contradicts the finding that contrastive learning requires large batch sizes~\cite{chen2020simple}. It is probably due to that all SimCSE models start from pre-trained checkpoints, which already provide us a good set of initial parameters.

\begin{table}[h]
    \begin{center}
    \centering
    \resizebox{0.98\columnwidth}{!}{%
    \begin{tabular}{lcccccc}
    \toprule
    & \multicolumn{4}{c}{\tf{Unsupervised}} &  \multicolumn{2}{c}{\tf{Supervised}} \\
    & \multicolumn{2}{c}{BERT} & \multicolumn{2}{c}{RoBERTa}  & \multirow{2}{*}{base} & \multirow{2}{*}{large} \\
    & base & large & base & large \\
    \midrule
    Batch size & 64 & 64 & 512 & 512 & 512& 512\\
    Learning rate & 3e-5 & 1e-5 & 1e-5 & 3e-5 & 5e-5 & 1e-5\\
    \bottomrule
    \end{tabular}
    }
    \end{center}
    \caption{
        Batch sizes and learning rates for SimCSE.
    }
    \label{tab:hyper}
\end{table}

For both unsupervised and supervised SimCSE, we take the \cls~representation with an MLP layer on top of it as the sentence representation. Specially, for unsupervised SimCSE, we discard the MLP layer and only use the \cls~output during test, since we find that it leads to better performance (ablation study in \S\ref{sec:ablation}).

Finally, we introduce one more optional variant which adds a masked language modeling (MLM) objective~\cite{devlin-etal-2019-bert} as an auxiliary loss to Eq.~\ref{eq:objective}: $\ell + \lambda \cdot \ell^{\mr{mlm}}$ ($\lambda$ is a hyperparameter). This helps {\ours} avoid catastrophic forgetting of token-level knowledge. As we will show in Table~\ref{tab:ablation}, we find that adding this term can help improve performance on transfer tasks (not on sentence-level STS tasks).

\section{Different Settings for STS Evaluation}
\label{sec:sts_discrepancy}


\begin{table}[t]
    \begin{center}
    \centering
    \resizebox{0.98\columnwidth}{!}{%
    \begin{tabular}{lccc}
    \toprule
        Paper & Reg. &  Metric & Aggr. \\
    \midrule
        \citet{hill-etal-2016-learning} &  & Both & all \\
        \citet{conneau-etal-2017-supervised-infersent} &  \checkmark &  Pearson   & mean  \\
        \citet{conneau-kiela-2018-senteval} &  \checkmark &  Pearson   & mean  \\
        \citet{reimers-gurevych-2019-sentence} &  & Spearman & all \\
        \citet{zhang-etal-2020-unsupervised} & & Spearman & all \\
        \citet{li-etal-2020-sentence} &  & Spearman & wmean \\
        \citet{su2021whitening} & &  Spearman & wmean\\
        \citet{wieting-etal-2020-bilingual} && Pearson & mean \\
        \citet{giorgi2020declutr} && Spearman & mean \\
        Ours &  & Spearman & all \\
    \bottomrule
    \end{tabular}
    }
    \end{center}

    \caption{
        STS evaluation protocols used in different papers. ``Reg.'': whether an additional regressor is used; ``aggr.'': methods to aggregate different subset results.
    }
    \label{tab:metric_used}
\end{table}

We elaborate the differences in STS evaluation settings in previous work in terms of (a) whether to use additional regressors; (b) reported metrics; (c) different ways to aggregate results.


\begin{table*}[t]
    \begin{center}
    \centering
    \small

    \begin{tabular}{lcccccccc}
    \toprule
       \tf{Model} & \tf{STS12} & \tf{STS13} & \tf{STS14} & \tf{STS15} & \tf{STS16} & \tf{STS-B} & \tf{SICK-R} & \tf{Avg.} \\
    \midrule
        SBERT (all)& 70.97&	76.53&	73.19&	79.09&	74.30&	76.98&	72.91&	74.85\\
        SBERT (wmean) & 66.35 & 73.76 & 73.88 & 77.33 & 73.62 & 76.98 & 72.91 & 73.55\\
        SBERT$^\clubsuit$ & 70.97 & 76.53 & 73.19 & 79.09 & 74.30 & 77.03 & 72.91 & 74.89\\
    \midrule
        BERT-whitening (NLI, all) &57.83 & 66.90 & 60.89 & 75.08 & 71.30 & 68.23 & 63.73 & 66.28\\
        BERT-whitening (NLI, wmean) & 61.43 & 65.90 & 65.96 & 74.80 & 73.10 & 68.23 & 63.73 & 67.59\\
        BERT-whitening (NLI)$^\spadesuit$ & 61.69 & 65.70 & 66.02 & 75.11 & 73.11 & 68.19 & 63.60 & 67.63\\
        BERT-whitening (target, all) & 42.88&	77.77&	66.27&	63.60&	67.58&	71.34&	60.40&	64.26\\
        BERT-whitening (target, wmean) & 63.38 & 73.01 & 69.13 & 74.48 & 72.56 & 71.34 & 60.40 & 69.19\\
        BERT-whitening (target)$^\spadesuit$ & 63.62 & 73.02 & 69.23 & 74.52 & 72.15 & 71.34 & 60.60 & 69.21\\
    \bottomrule
    \end{tabular}
    \end{center}

    \caption{
       Comparisons of our reproduced results using different evaluation protocols and the original numbers.
       $\clubsuit$: results from \citet{reimers-gurevych-2019-sentence};
       $\spadesuit$: results from \citet{su2021whitening};
       Other results are reproduced by us.
        From the table we see that SBERT takes the ``all'' evaluation and BERT-whitening takes the ``wmean'' evaluation.
    }
    \label{tab:replicated_results}
\end{table*}


\begin{table*}[t]
    \begin{center}
    \centering
    \small
    
    \begin{tabular}{lcccccccc}
    \toprule
       \tf{Model} & \tf{STS12} & \tf{STS13} & \tf{STS14} & \tf{STS15} & \tf{STS16} & \tf{STS-B} & \tf{SICK-R} & \tf{Avg.} \\
    \midrule
        BERT\ba~(first-last avg.)$^\spadesuit$ & 57.86 & 61.97 &  62.49 & 70.96 & 69.76 & 59.04 & 63.75 & 63.69\\
        \tableindent + flow (NLI)$^\spadesuit$ & 59.54 & 64.69 & 64.66 & 72.92 & 71.84 & 58.56 & 65.44 & 65.38\\ 
        \tableindent + flow (target)$^\spadesuit$ & 63.48 & 72.14 & 68.42 & 73.77 & 75.37 & 70.72 & 63.11 & 69.57\\ 
        \tableindent + whitening (NLI)$^\spadesuit$ & 61.69 & 65.70 & 66.02 & 75.11 & 73.11 & 68.19 & 63.60 & 67.63\\
        \tableindent + whitening (target)$^\spadesuit$ & 63.62 & 73.02 & 69.23 & 74.52 & 72.15 & 71.34 & 60.60 & 69.21\\ 
        $*$ Unsup. \ours-BERT\ba & \tf{70.14} & \tf{79.56} & \tf{75.91} & \tf{81.46} & \tf{79.07} &\tf{76.85}	& \tf{72.23} & \tf{76.46} \\
        \midrule
        SBERT\ba~(first-last avg.)$^\spadesuit$ & 68.70 & 74.37 & 74.73 & 79.65 & 75.21 & 77.63 &   74.84 & 75.02\\
        \tableindent + flow (NLI)$^\spadesuit$ & 67.75 & 76.73 & 75.53 & 80.63 & 77.58 & 79.10 & 78.03 & 76.48\\ 
        \tableindent + flow (target)$^\spadesuit$ & 68.95 & 78.48 & 77.62 & 81.95 & 78.94 & 81.03 & 74.97 & 77.42 \\
        \tableindent + whitening (NLI)$^\spadesuit$ & 69.11 & 75.79 & 75.76 & 82.31 & 79.61 & 78.66 & 76.33 & 76.80\\
        \tableindent + whitening (target)$^\spadesuit$ & 69.01 & 78.10 & 77.04 & 80.83 & 77.93 & 80.50 & 72.54 & 76.56\\ 
        $*$ Sup. \ours-BERT\ba & \tf{70.90}&	\tf{81.49}&	\tf{80.19}&	\tf{83.79}&	\tf{81.89}&	\tf{84.25}&	\tf{80.39}&	\tf{80.41} \\
    \bottomrule
    \end{tabular}
    \end{center}

    \caption{
        STS results with ``wmean'' setting (Spearman). 
        $\spadesuit$: from \citet{li-etal-2020-sentence,su2021whitening}.
    }
    \label{tab:wmean_sts}
\end{table*}

\tf{Additional regressors.} The default SentEval implementation applies a linear regressor on top of frozen sentence embeddings for STS-B and SICK-R, and train the regressor on the training sets of the two tasks, while most sentence representation papers take the raw embeddings and evaluate in an unsupervised way. In our experiments, \emph{we do not apply any additional regressors and directly take cosine similarities for all STS tasks.}

\tf{Metrics.} Both Pearson's and Spearman's correlation coefficients are used in the literature. \citet{reimers-etal-2016-task} argue that Spearman correlation,
which measures the rankings instead of the actual scores,
better suits the need of evaluating sentence embeddings. \emph{For all of our experiments, we report Spearman's rank correlation.}

\tf{Aggregation methods.} Given that each year's STS challenge contains several subsets, there are different choices to gather results from them: one way is to concatenate all the topics and report the overall Spearman's correlation (denoted as ``all''), and the other is to calculate results for different subsets separately and average them (denoted as ``mean'' if it is simple average or ``wmean'' if weighted by the subset sizes). However, most papers do not claim the method they take, making it challenging for a fair comparison.
We take some of the most recent work: SBERT~\cite{reimers-gurevych-2019-sentence}, BERT-flow~\cite{li-etal-2020-sentence} and BERT-whitening~\cite{su2021whitening}\footnote{\citet{li-etal-2020-sentence} and \citet{su2021whitening} have consistent results, so we assume that they take the same evaluation and just take BERT-whitening in experiments here.} as an example:
In Table~\ref{tab:replicated_results}, we compare our reproduced results to reported results of SBERT and BERT-whitening, and  find that \citet{reimers-gurevych-2019-sentence} take the ``all'' setting but \citet{li-etal-2020-sentence,su2021whitening} take the ``wmean'' setting, even though \citet{li-etal-2020-sentence} claim that they take the same setting as \citet{reimers-gurevych-2019-sentence}.
Since the ``all'' setting fuses data from different topics together, it makes the evaluation closer to real-world scenarios, and unless specified, \emph{we take the ``all'' setting.}

We list evaluation settings for a number of previous work in Table~\ref{tab:metric_used}.
Some of the settings are reported by the paper and some of them are inferred by comparing the results and checking their code.
As we can see, the evaluation protocols are very incoherent across different papers.
We call for unifying the setting in evaluating sentence embeddings for future research.
We will also release our evaluation code for better reproducibility.
Since previous work uses different evaluation protocols from ours, we further evaluate our models in these settings to make a direct comparison to the published numbers.
We evaluate \ours with ``wmean'' and Spearman's correlation to directly compare to
\citet{li-etal-2020-sentence} and \citet{su2021whitening} in Table~\ref{tab:wmean_sts}.

\section{Baseline Models}
\label{sec:baseline}


\begin{table*}[ht]
    \begin{center}
    \centering
    \resizebox{0.9\textwidth}{!}{%

    \begin{tabular}{lcccccccc}
    \toprule
       \tf{Model} & \tf{STS12} & \tf{STS13} & \tf{STS14} & \tf{STS15} & \tf{STS16} & \tf{STS-B} & \tf{SICK-R} & \tf{Avg.} \\
       \midrule
        BERT-flow (NLI) & 58.40&	67.10&	60.85&	75.16&	71.22&	68.66&	64.47&	66.55\\
        BERT-flow (target) & 53.15&	78.38&	66.02&	62.09&	70.84&	71.70 &	61.97&	66.31\\
        BERT-whitening (NLI) & 57.83& 66.90 & 60.90 & 75.08& 71.31& 68.24& 63.73& 66.28\\
        BERT-whitening (target) & 42.88&	77.77&	66.28&	63.60&	67.58&	71.34&	60.40&	64.26\\
    \midrule
        SBERT-flow (NLI) & 69.78&	77.27&	74.35&	82.01&	77.46&	79.12&	76.21&	76.60\\
        SBERT-flow (target) & 66.18&	82.69&	76.22&	73.72&	75.71&	79.99&	73.82&	75.48\\
        SBERT-whitening (NLI) & 69.65&	77.57&	74.66&	82.27&	78.39&	79.52&	76.91&	77.00\\
        SBERT-whitening (target) & 52.91&	81.91&	75.44&	72.24&	72.93&	80.50&	72.54&	72.64\\
    \bottomrule
    \end{tabular}
    }
    \end{center}

    \caption{
        Comparison of using NLI or target data for postprocessing methods (``all'', Spearman's correlation).
    }
    \label{tab:target_vs_nli}
\end{table*}

We elaborate on how we obtain different baselines for comparison in our experiments:
\begin{itemize}
    \item For average GloVe embedding \cite{pennington-etal-2014-glove}, InferSent \cite{conneau-etal-2017-supervised-infersent} and Universal Sentence Encoder \cite{cer-etal-2018-universal}, we directly report the results from \citet{reimers-gurevych-2019-sentence}, since our evaluation setting is the same as theirs.
    \item For BERT \cite{devlin-etal-2019-bert} and RoBERTa \cite{liu2019roberta}, we download the pre-trained model weights from HuggingFace's \texttt{Transformers}\footnote{\url{https://github.com/huggingface/transformers}}, and evaluate the models with our own scripts.
    \item For SBERT and SRoBERTa \cite{reimers-gurevych-2019-sentence}, we reuse the results from the original paper. For results not reported by \citet{reimers-gurevych-2019-sentence}, such as the performance of SRoBERTa on transfer tasks, we download the model weights from SentenceTransformers\footnote{\url{https://www.sbert.net/}} and evaluate them.
    \item
    For DeCLUTR~\cite{giorgi2020declutr} and contrastive tension~\cite{carlsson2021semantic}, we reevaluate their  checkpoints in our setting.
    \item
    For BERT-flow~\cite{li-etal-2020-sentence}, since their original numbers take a different setting, we retrain their models using their code\footnote{\url{https://github.com/bohanli/BERT-flow}}, and evaluate the models using our own script.
    \item For BERT-whitening \cite{su2021whitening}, we implemented our own version of whitening script following the same pooling method in \citet{su2021whitening}, i.e. first-last average pooling. Our implementation can reproduce the results from the original paper (see Table~\ref{tab:replicated_results}).
\end{itemize}


For both BERT-flow and BERT-whitening, they have two  variants of postprocessing: one takes the NLI data (``NLI'') and one directly learns the embedding distribution on the target sets (``target'').
We find that in our evaluation setting, ``target'' is generally worse than ``NLI'' (Table~\ref{tab:target_vs_nli}), so we only report the NLI variant in the main results.

\section{Ablation Studies}

\label{sec:app_ablation}






\begin{table}[t]
    \begin{center}
    \centering
    \resizebox{0.9\columnwidth}{!}{%
    \begin{tabular}{lcccccc}
    \toprule
        $\tau$ & N/A & 0.001 & 0.01 & 0.05 & 0.1 & 1\\
    \midrule
        \tf{STS-B} & 85.9 & 84.9 & 85.4 & \tf{86.2} & 82.0 & 64.0 \\
    \bottomrule
    \end{tabular}
    }
    \end{center}

    \caption{
        STS-B development results (Spearman's correlation) with different temperatures. ``N/A'': Dot product instead of cosine similarity.
    }
    \label{tab:temperature}
\end{table}

\paragraph{Normalization and temperature.}
We train \ours using both dot product and cosine similarity with different temperatures and evaluate them on the STS-B development set. As shown in Table~\ref{tab:temperature},
with a carefully tuned temperature $\tau = 0.05$, cosine similarity is better than dot product.

\paragraph{MLM auxiliary task.} Finally, we study the impact of the MLM auxiliary objective with different $\lambda$. As shown in Table~\ref{tab:ablation}, the token-level MLM objective improves the averaged performance on transfer tasks modestly, yet it brings a consistent drop in semantic textual similarity tasks.


\begin{table}[t]
    \begin{center}
    \centering
    \small
    \resizebox{0.8\columnwidth}{!}{%
    \begin{tabular}{lcc}
    \toprule
        \tf{Model} & \tf{STS-B} & \tf{Avg. transfer} \\
    \midrule
        w/o MLM &  \tf{86.2} & 85.8 \\
        w/ MLM \\
        \tableindent $\lambda=0.01$ & 85.7 & 86.1\\
        \tableindent $\lambda=0.1$ & 85.7 & \tf{86.2}\\
        \tableindent $\lambda=1$ & 85.1 & 85.8 \\
    \bottomrule
    \end{tabular}
    }
    \end{center}

    \caption{
        Ablation studies of the MLM objective based on the development sets using BERT\ba.
    }
    \label{tab:ablation}
\end{table}

\section{Transfer Tasks}
\label{sec:transfer}


\begin{table*}[ht]
    \begin{center}
    \centering
    \small

    \begin{tabular}{lcccccccc}
    \toprule
       \tf{Model} & \tf{MR} & \tf{CR} & \tf{SUBJ} & \tf{MPQA} & \tf{SST} & \tf{TREC} & \tf{MRPC} & \tf{Avg.}\\
    \midrule
    \midrule
        \multicolumn{9}{c}{\it{Unsupervised models}}\\
    \midrule
        GloVe embeddings (avg.)$^\clubsuit$ & 77.25&    78.30&  91.17&  87.85&  80.18&  83.00& 72.87 & 81.52\\
        Skip-thought$^\heartsuit$ &  76.50& 80.10&  93.60&  87.10&  82.00&  92.20&  73.00& 83.50  \\
        \midrule
        Avg. BERT embeddings$^\clubsuit$ & 78.66 & 86.25 & 94.37 & 88.66 & 84.40 & \tf{92.80} & 69.54 & 84.94 \\
        BERT-\cls embedding$^\clubsuit$ & 78.68 & 84.85 & 94.21 & 88.23 & 84.13 & 91.40 & 71.13 & 84.66 \\
        IS-BERT\ba$^\heartsuit$ & 81.09 & 87.18 & 94.96 & 88.75 & 85.96 & 88.64 & 74.24 & 85.83 \\
        $*$ \ours-BERT\ba & 81.18&	86.46&	94.45&	\tf{88.88}&	85.50&	89.80&	74.43&	85.81\\
        \tableindent w/ MLM & \textbf{82.92}&	\textbf{87.23}&	\tf{95.71}&	88.73&	\tf{86.81}&	87.01&	\tf{78.07}& \tf{86.64}\\
        \midrule
        $*$ \ours-RoBERTa\ba & 81.04 &	87.74 &	93.28 &	86.94	&86.60	&84.60	&73.68 &84.84 \\
        \tableindent w/ MLM & \tf{83.37}& 	\tf{87.76}& 	\tf{95.05}	& \tf{87.16}	& \tf{89.02}	& \tf{90.80}	& \tf{75.13} & \tf{86.90}\\
        $*$ \ours-RoBERTa\la & 82.74 &	87.87&	93.66&	\tf{88.22}	&88.58	&92.00	&69.68 & 86.11 \\
        \tableindent w/ MLM & \tf{84.66}	&\tf{88.56}&	\tf{95.43}&	87.50&	\tf{89.46}&	\tf{95.00}&	\tf{72.41} & \tf{87.57} \\
    \midrule
    \midrule
        \multicolumn{9}{c}{\it{Supervised models}}\\
    \midrule
        InferSent-GloVe$^\clubsuit$ & 81.57 & 86.54 & 92.50 & {90.38} & 84.18 & 88.20 & 75.77 & 85.59\\
        Universal Sentence Encoder$^\clubsuit$ & 80.09 & 85.19 & 93.98 & 86.70 & 86.38 & {93.20} & 70.14 & 85.10  \\
        \midrule
        SBERT\ba$^\clubsuit$ & \tf{83.64}&	\tf{89.43}&	94.39&	\tf{89.86} &	\tf{88.96}&	\tf{89.60}&	76.00&	\tf{87.41}\\ 
        $*$ \ours-BERT\ba  & 82.69&	89.25&	\tf{94.81}&	89.59&	87.31&	88.40&	73.51&	86.51\\
        \tableindent w/ MLM & 82.68&	88.88&	94.52&	89.82&	88.41&	87.60&	\tf{76.12}&	86.86 \\
        \midrule
        SRoBERTa\ba & 84.91&	90.83&	92.56&	88.75&	90.50&	88.60&	\tf{78.14}&	87.76\\ 
        $*$ \ours-RoBERTa\ba  & 84.92&	\tf{92.00}&	\tf{94.11}&	\tf{89.82}&	91.27&	88.80&	75.65&	88.08\\ 
        \tableindent w/ MLM & \tf{85.08}&	91.76&	94.02&	89.72&	\tf{92.31}&	\tf{91.20}&	76.52&	\tf{88.66}\\
        $*$ \ours-RoBERTa$_{\texttt{large}}$ & 88.12&	92.37&	95.11&	90.49&	92.75&	91.80&	76.64&	89.61\\
        \tableindent w/ MLM & \tf{88.45}&	\tf{92.53}&	\tf{95.19}&	\tf{90.58}&	\tf{93.30}&	\tf{93.80}&	\tf{77.74}&	\tf{90.23} \\
    \bottomrule
    \end{tabular}
    \end{center}

    \caption{
        Transfer task results of different sentence embedding models (measured as accuracy). $\clubsuit$: results from \citet{reimers-gurevych-2019-sentence};
        $\heartsuit$: results from \citet{zhang-etal-2020-unsupervised}. We highlight the highest numbers among models with the same pre-trained encoder. MLM: adding MLM as an auxiliary task with $\lambda = 0.1$. 
    }
    \label{tab:main_transfer}
\end{table*}

We  evaluate our models on the following transfer tasks: MR~\cite{pang2005seeing_mr}, CR~\cite{hu2004mining_cr}, SUBJ~\cite{pang2004sentimental_subj}, MPQA~\cite{wiebe2005annotating_mpqa}, SST-2~\cite{socher2013recursive_sst-2}, TREC~\cite{voorhees2000building_trec} and MRPC~\cite{dolan-brockett-2005-automatically-mrpc}.
A logistic regression classifier is trained on top of (frozen) sentence embeddings produced by different methods. We follow default configurations from SentEval\footnote{\url{https://github.com/facebookresearch/SentEval}}.

Table~\ref{tab:main_transfer} shows the evaluation results on transfer tasks.
We find that supervised {\ours} performs on par or better than previous approaches, although the trend of unsupervised models remains unclear. We find that adding this MLM term consistently improves performance on transfer tasks, confirming our intuition that sentence-level objective may not directly benefit transfer tasks.  We also experiment with post-processing methods (BERT-flow/whitening) and find that they both hurt performance compared to their base models, showing that good uniformity of representations does not lead to better embeddings for transfer learning. As we argued earlier, we think that transfer tasks are not a major goal for sentence embeddings, and thus we take the STS results for main comparison.

\section{Distribution of Singular Values}
\label{sec:singular_value}

\begin{figure}
    \begin{minipage}[t]{0.49\linewidth}
    \centering
    \centerline{\includegraphics[width=\textwidth]{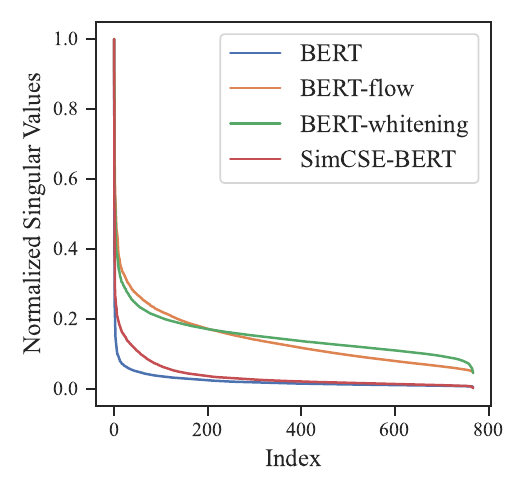}}
    \centerline{\small Unsupervised models}
    \end{minipage}
    \begin{minipage}[t]{0.49\linewidth}
    \centering
    \includegraphics[width=\textwidth]{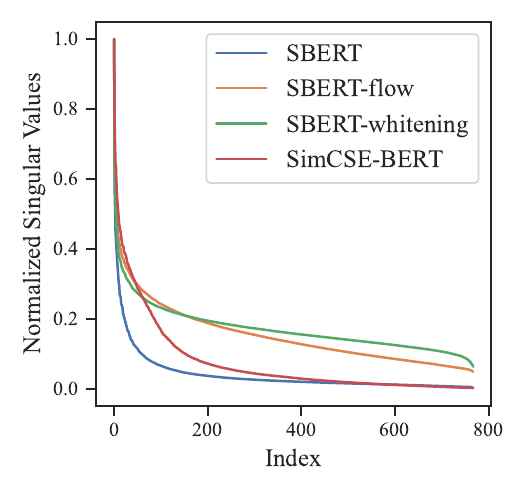}
    \centerline{\small Supervised models}
    \end{minipage}
    \centering
    \caption{Singular value distributions of sentence embedding matrix from sentences in STS-B. We normalize the singular values so that the largest one is 1.}
    \label{fig:singular}
    \vspace{-4pt}
\end{figure}

Figure~\ref{fig:singular} shows the singular value distribution of \ours together with other baselines. For both unsupervised and supervised cases, singular value drops the fastest for vanilla BERT or SBERT embeddings, while \ours helps flatten the spectrum distribution. Postprocessing-based methods such as BERT-flow or BERT-whitening flatten the curve even more since they directly aim for the goal of mapping embeddings to an isotropic distribution.

\section{Cosine-similarity Distribution}
\label{sec:cos-sim-dis}


\begin{figure*}[t]
    \label{denseplot}
    \centering

    \begin{minipage}[t]{0.31\linewidth}
    \centering
    \centerline{\includegraphics[width=\textwidth]{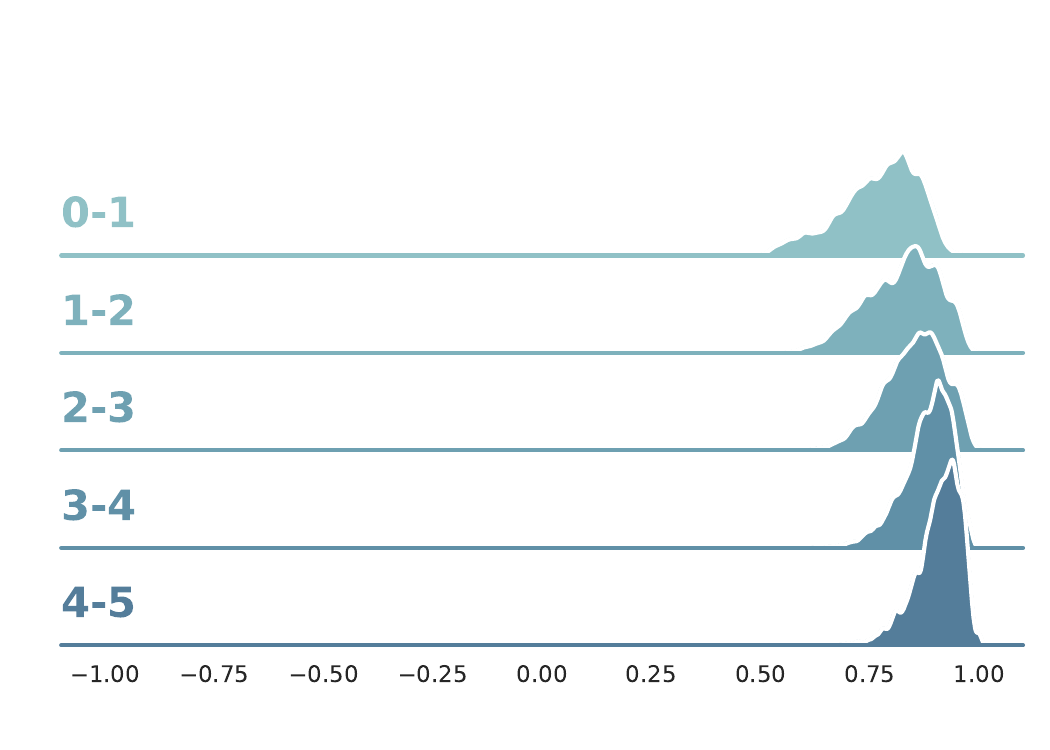}}
    \centerline{\small Avg. BERT\ba}
    \end{minipage}
    \begin{minipage}[t]{0.31\linewidth}
    \centering
    \includegraphics[width=\textwidth]{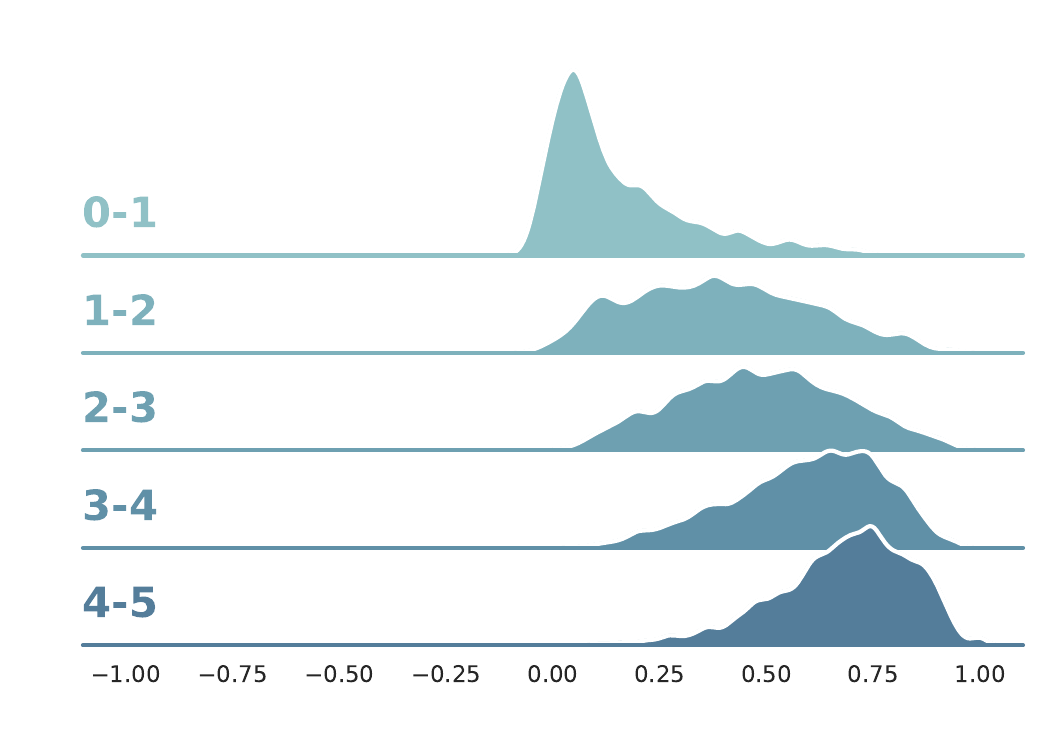}
    \centerline{\small BERT\ba-whitening}
    \end{minipage}
    \begin{minipage}[t]{0.31\linewidth}
    \centering
    \includegraphics[width=\textwidth]{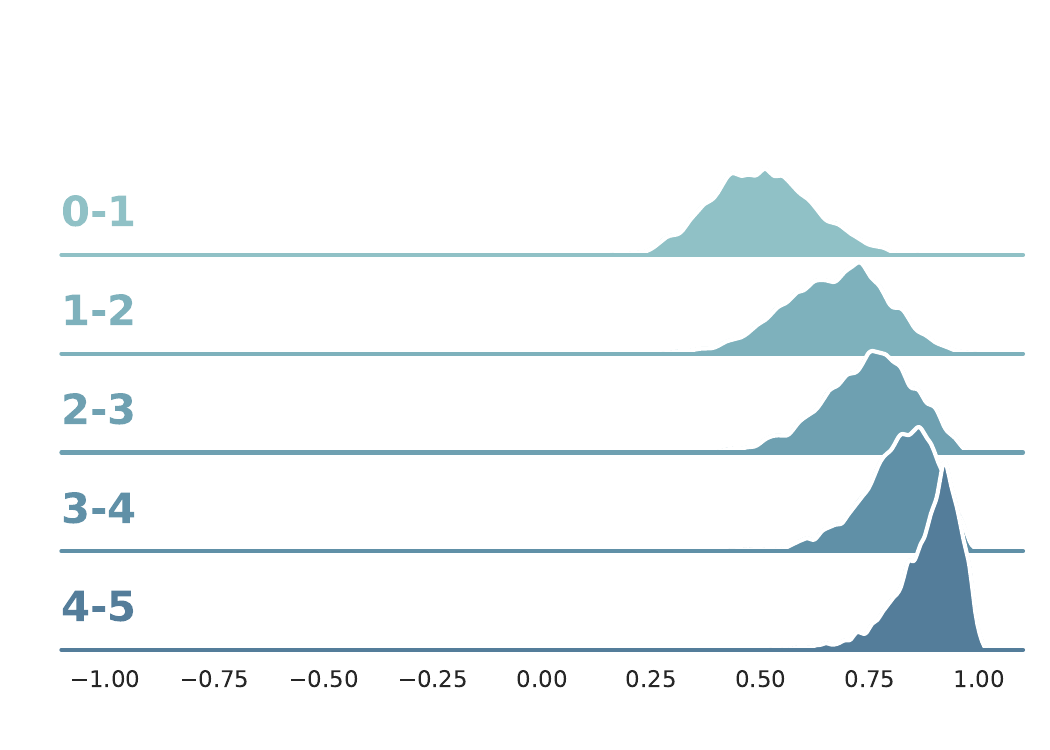}
    \centerline{\small Unsupervised \ours-BERT\ba}
    \end{minipage}

    \begin{minipage}[t]{0.31\linewidth}
    \centering
    \centerline{\includegraphics[width=\textwidth]{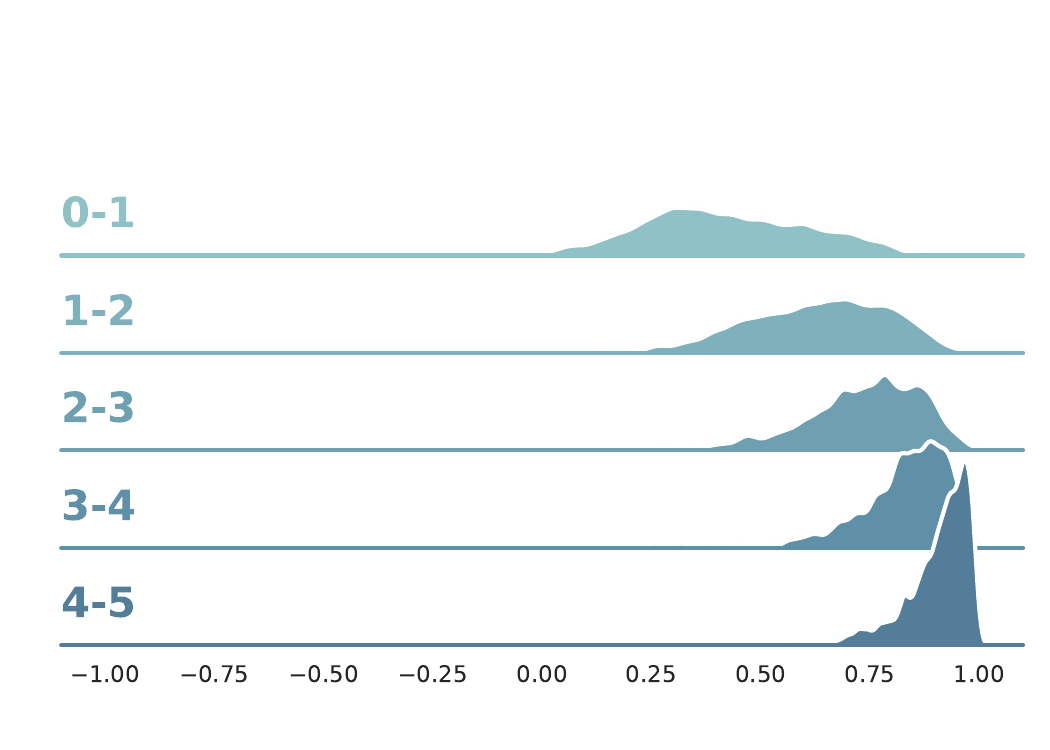}}
    \centerline{\small SBERT\ba}
    \end{minipage}
    \begin{minipage}[t]{0.31\linewidth}
    \centering
    \includegraphics[width=\textwidth]{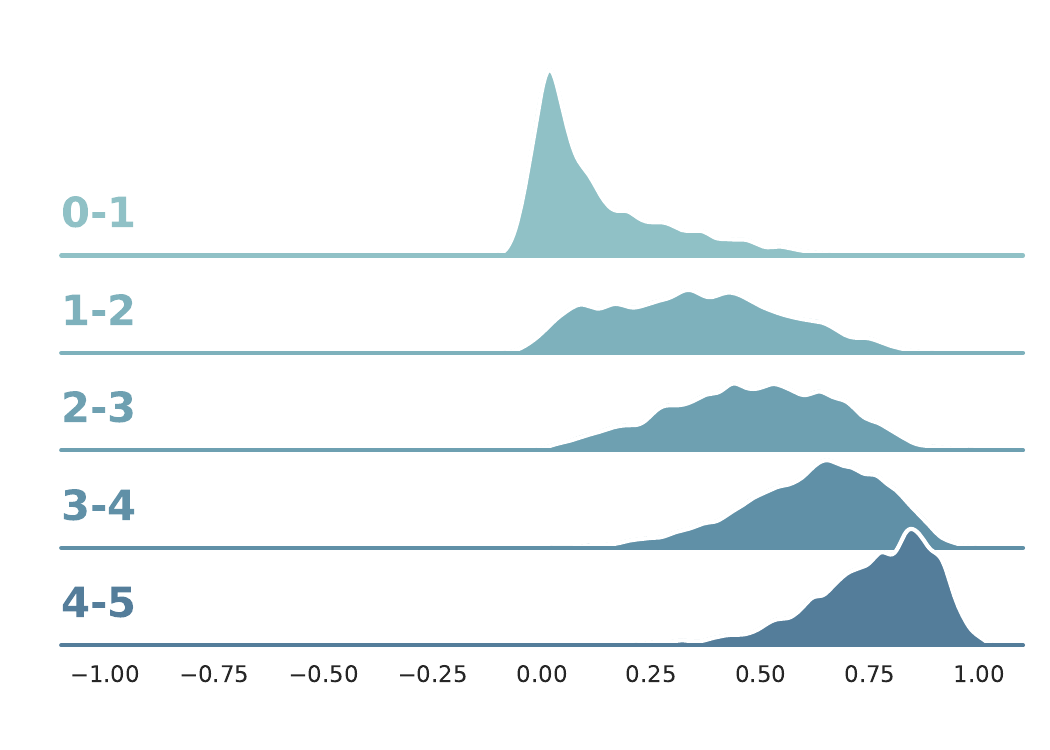}
    \centerline{\small SBERT\ba-whitening}
    \end{minipage}
    \begin{minipage}[t]{0.31\linewidth}
    \centering
    \includegraphics[width=\textwidth]{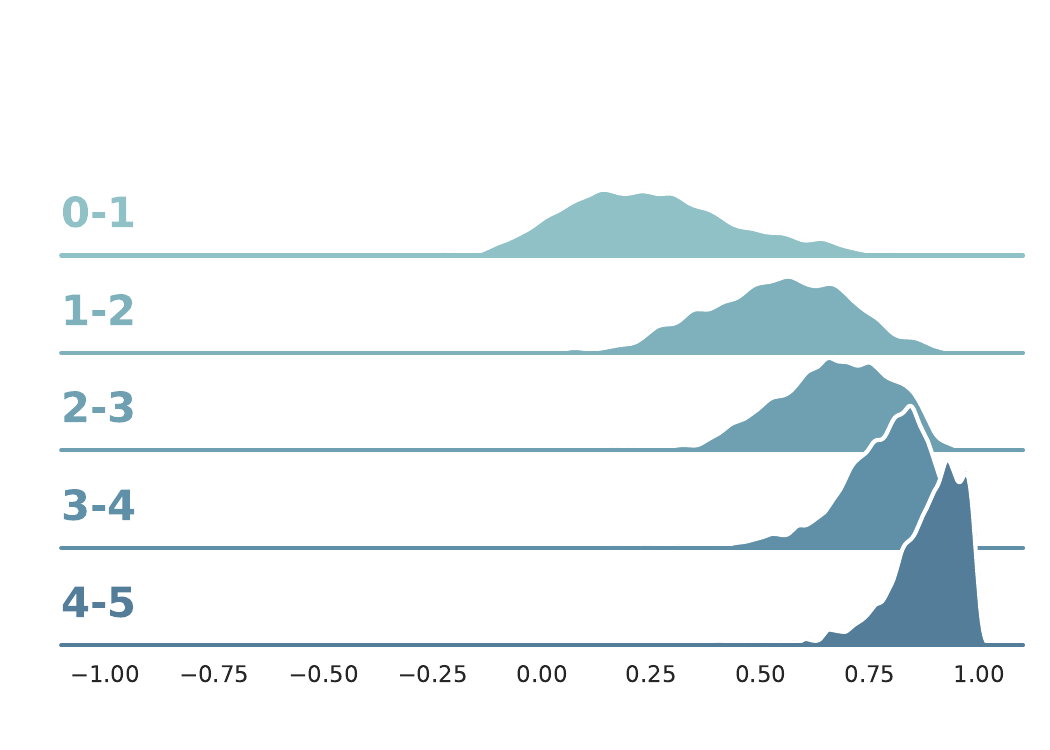}
    \centerline{\small Supervised \ours-BERT\ba}
    \end{minipage}

    \centering
    \caption{Density plots of cosine similarities between sentence pairs in STS-B. Pairs are divided into 5 groups based on ground truth ratings (higher means more similar) along the y-axis, and x-axis is the cosine similarity.} 
    \vspace{6pt}
    \label{fig:histogram}
    \end{figure*}

To directly show the strengths of our approaches on STS tasks, we illustrate the cosine similarity distributions of STS-B pairs with different groups of human ratings in
Figure~\ref{fig:histogram}.
Compared to all the baseline models, both unsupervised and supervised \ours better distinguish sentence pairs with different levels of similarities, thus leading to a better performance on STS tasks. In addition, we observe that \ours generally shows a more scattered distribution than BERT or SBERT, but also preserves a lower variance on semantically similar sentence pairs compared to whitened distribution. This observation further validates that \ours can achieve a better alignment-uniformity balance.

\end{document}